\documentclass{article}
\ifdefined\pdfobjcompresslevel
\fi
\ifdefined\pdfvariable
  \pdfvariable objcompresslevel=0
\fi

\usepackage[preprint, eandd]{neurips_2026}

\usepackage[utf8]{inputenc}
\usepackage[T1]{fontenc}
\usepackage{hyperref}
\usepackage{url}
\usepackage{booktabs}
\usepackage{amsfonts}
\usepackage{amsmath}
\usepackage{nicefrac}
\usepackage{microtype}
\usepackage[table]{xcolor}
\usepackage{graphicx}
\usepackage{enumitem}
\usepackage{array}
\usepackage{multirow}
\usepackage{tabularx}
\usepackage{longtable}

\title{Video-HOCA: A Diagnostic Benchmark for Physical Anomaly Reasoning in Video-LLMs}

\author{%
  Chang Liu$^{1}$\quad
  Yunfan Ye$^{2}$\quad
  Qingyang Zhou$^{1}$\quad
  Xichen Tan$^{1}$\quad
  Mengxuan Luo$^{1}$ \\[0.4em]
  \bfseries
  Zhenyu Qiu$^{1}$\quad
  Wei Peng$^{1}$\quad
  Zhiping Cai$^{1,\ast}$ \\[0.6em]
  \normalfont\small
  $^{1}$National University of Defense Technology, Changsha, China \quad
  $^{2}$Hunan University, Changsha, China \\[0.2em]
  $^{\ast}$Corresponding author: \texttt{zpcai@nudt.edu.cn}
}

\begin{document}

\maketitle

\begin{abstract}
    We introduce Video-HOCA, a diagnostic benchmark for physical anomaly
    reasoning in videos. Video-HOCA uses an Ontological--Causal taxonomy to
    distinguish violations of an entity's own properties or capabilities from
    violations of physical relations among entities and the environment. It
    contains more than 1{,}400 generated and real-world videos and 3{,}470
    question-answer pairs, with human verification of labels and reference
    answers. The benchmark evaluates four levels of reasoning: plausibility
    checking, anomaly attribution, fine-grained recognition, and open-ended
    physical explanation. Across 20 Instruct-mode Video-LLMs, we find that
    recognition outpaces explanation: Task~I scores cluster at $75$--$88$,
    while Task~II macro-F1 stays mostly below $50$. We also find that the
    Ontological--Causal gap depends on the task and model configuration, and
    that Thinking-mode gains are not explained by sampling or output budget
    alone. Annotation agreement, Task-IV human--judge and judge--judge checks,
    alternative metrics, and temporal/decoding controls validate the evaluation
    pipeline and bound the claims supported by the benchmark.
\end{abstract}

\section{Introduction}

Video-language models can now describe many visible events in videos, but
description is not the same as physical reasoning. A model may correctly name
the objects in a scene and still miss why the event is physically impossible.
This distinction matters for physically anomalous videos: the question is not
only whether a model sees something unusual, but whether it can identify the
violated physical regularity instead of relying on surface patterns
\citep{lecun2022path}.

Existing video benchmarks only partially test this ability. General video
benchmarks often emphasize recognition, temporal retrieval, or knowledge-heavy
question answering. Recent hallucination and impossible-video benchmarks move
closer to physical plausibility, but many of them still focus on binary realism
checks, generic impossible events, or generation faithfulness. They do not
systematically separate several evaluation levels that are easy to conflate:
detecting an implausible video, attributing the anomaly to a physical category,
recognizing the specific anomalous event, and explaining the physical mechanism
behind it. Without this separation, a model that recognizes an anomaly can look
more physically grounded than it really is.

We introduce \textbf{Video-HOCA}, a diagnostic benchmark for physical anomaly
reasoning in videos. Video-HOCA uses an operational taxonomy rather than a
philosophical one. The top-level split separates \textbf{Ontological} anomalies,
where the primary violation lies in an entity's own property, state, or
biological capability, from \textbf{Causal} anomalies, where otherwise valid
entities interact with each other or with the environment in a physically invalid
way. The benchmark contains generated anomaly videos and semantically matched
real videos, more than 1{,}400 videos in total, and 3{,}470 question-answer
pairs. Its four main tasks test plausibility checking, taxonomy attribution,
fine-grained anomaly recognition, and open-ended physical explanation.

Because Video-HOCA is meant to support evaluative claims, we validate the
evaluation pipeline instead of treating it as an implementation detail. We
measure re-annotation agreement for the taxonomy, validate the rubric-based
Task-IV judge with human scores and judge substitution, use macro-F1 for the
taxonomy-attribution task, check aggregation stability, and run controls for
temporal input and decoding settings. These checks are central to the benchmark:
they define which conclusions can be drawn and which would be over-interpretation.

Our contributions are threefold. (i)~We introduce \textbf{Video-HOCA}, a
diagnostic benchmark for physical anomaly reasoning that uses an operational
Ontological--Causal taxonomy, paired generated/real videos, and four tasks that
separate plausibility checking, taxonomy attribution, fine-grained recognition,
and open-ended physical explanation.
(ii)~We provide a validation protocol for the benchmark---re-annotation
agreement for the taxonomy, human--judge and judge--judge checks for
Task~IV, macro-F1 and aggregation checks, and controls for temporal input
and decoding---making evaluative claims testable rather than implicit.
(iii)~We use Video-HOCA to identify failure modes hidden by standard video QA
benchmarks: models often detect anomalies before they can attribute or explain
them, the Ontological--Causal gap is not uniformly one-sided, temporal demand
explains only part of the open-ended reasoning gap, and Thinking-mode
comparisons depend on decoding parameters and output budget.

\begin{figure}[t]
    \centering
    \includegraphics[width=\linewidth]{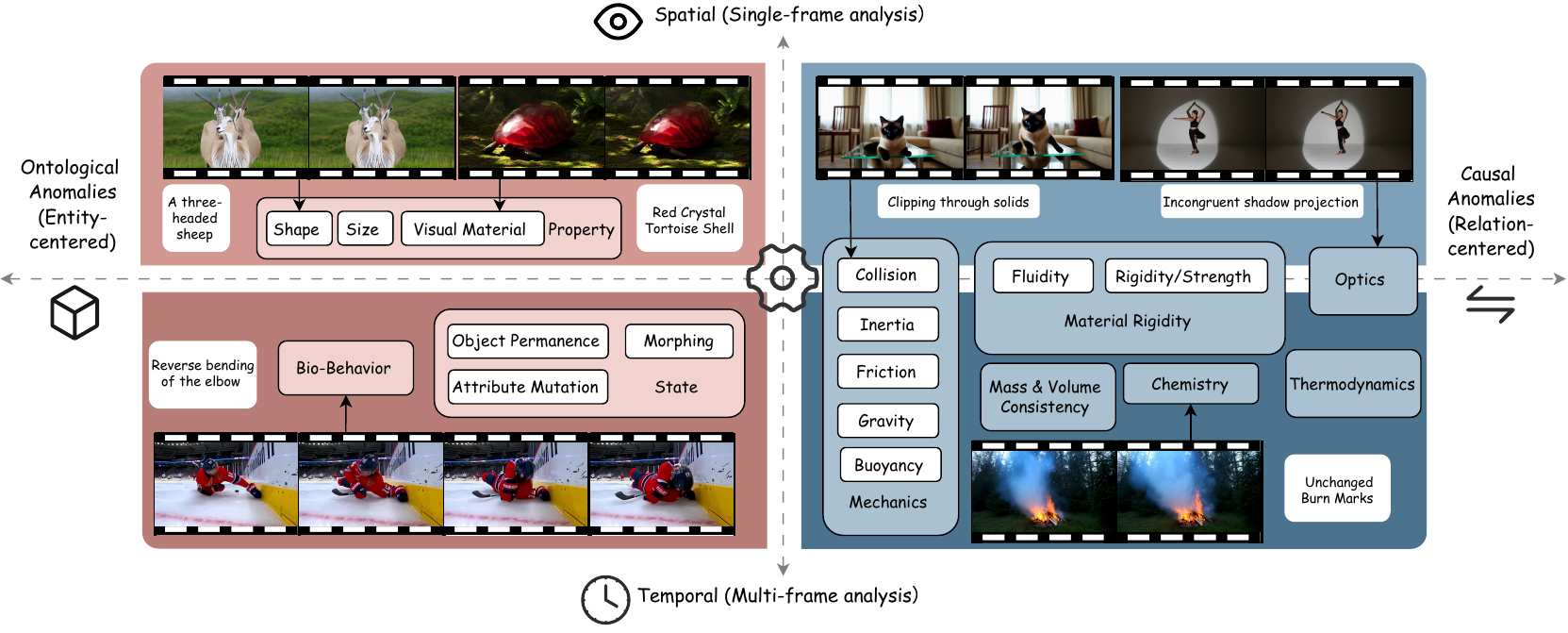}
    \caption{\textbf{Video-HOCA overview.} The benchmark organizes physical
             anomalies into Ontological (entity-centered) and Causal
             (relation-centered) branches, totaling 9 L2 groups and
             18 L3 tags (full hierarchy in Appendix
             Table~\ref{tab:taxonomy-def}); spatial and temporal cases span
             both generated anomaly videos and semantically matched real
             videos. The four main tasks move from plausibility checking to
             category attribution, fine-grained anomaly recognition, and
             open-ended physical explanation.} \label{fig:teaser}
    \vspace{-0.6em}
\end{figure}

\begin{figure}[t]
    \centering
    \includegraphics[width=\linewidth]{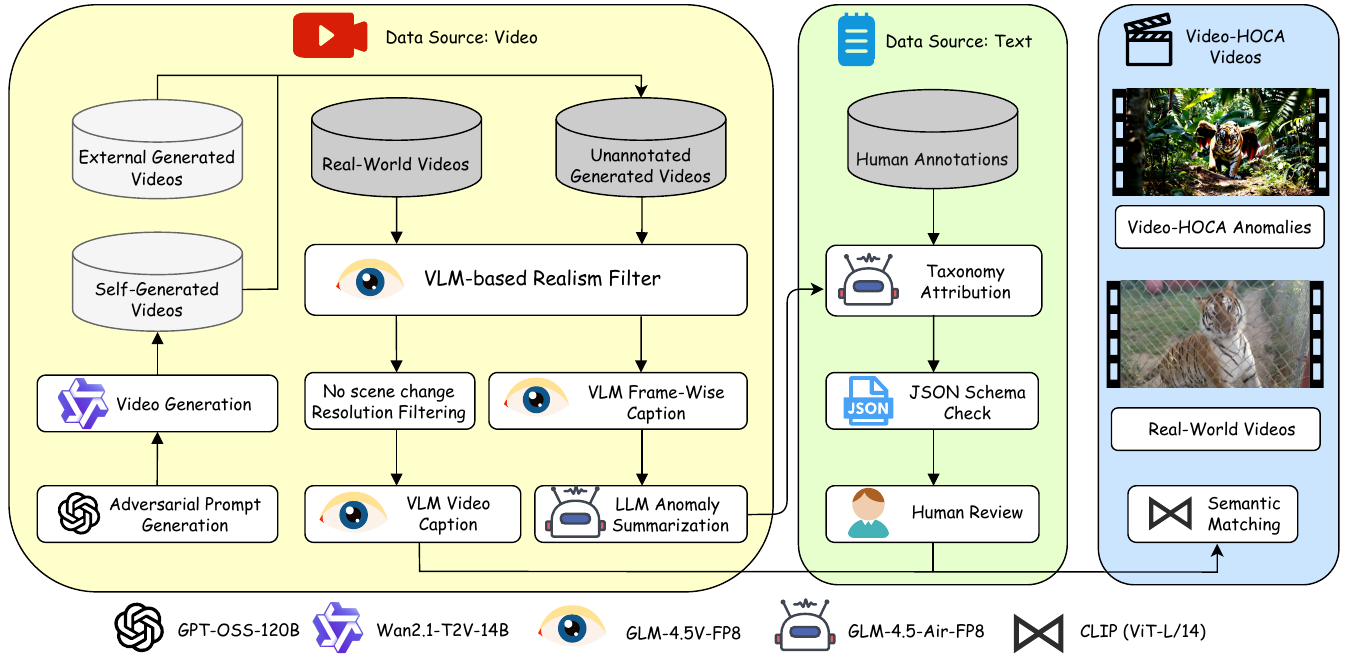}
\caption{\textbf{Construction pipeline.} We use both
             text-source candidates, where existing benchmarks provide videos
             with textual annotations, and video-source candidates, where raw
             generated clips require frame-wise captioning, anomaly
             summarization, operational taxonomy attribution, schema checking,
             and human review. Real-world clips are filtered and semantically
             matched to anomaly clips so that Task~I compares physically
             plausible and implausible videos under similar scene content.}
             \label{fig:pipeline}
    \vspace{-0.6em}
\end{figure}
\section{Related Work}

\paragraph{Video understanding benchmarks.}
Most video benchmarks test perception, temporal retrieval, or
knowledge-heavy question answering: Perception Test targets low-level
multimodal perception \citep{perceptiontest2023}, MVBench and MMBench-Video
cover broad video-understanding skills \citep{mvbench2024,mmbenchvideo2024},
Video-MME evaluates across video domains \citep{videomme2025}, and MMVU and
Video-MMMU add expert and discipline-level content
\citep{mmvu2025,videommmu2025}. These benchmarks mostly reward identifying
what is visible. Video-HOCA asks a narrower question: when a video is
physically implausible, can the model tell whether the problem lies in an
entity itself or in a physical relation among entities and the environment?
The Ontological--Causal split is loosely inspired by
work on object perception and early physical expectations
\citep{spelke1990principles,baillargeon2004infants}; we do not claim a
direct correspondence to those constructs and use the split purely as an
operational annotation rule for the dataset.

\paragraph{Physical plausibility, impossible videos, and video hallucination.}
Several recent benchmarks use physically implausible content as a probe:
VideoHallu studies multimodal hallucinations on synthetic videos
\citep{videohallu2025}; IPV-Bench builds impossible-video test cases
\citep{ipvbench2025}; VideoPhy-2 tests action-centric physical commonsense
\citep{videophy2_2025}; and VBench-2.0 studies generation faithfulness
\citep{vbench2_2025}. Video-HOCA uses the same idea but organizes anomalies
differently---separating Ontological and Causal violations, validating the
taxonomy by re-annotation, and progressing beyond plausibility
detection to category attribution, fine-grained recognition, and open-ended
physical explanation\citep{liu2025humansam,ye2024diffusionedge}.

\paragraph{Video-LLM model families and evaluation methodology.}
Modern Video-LLMs build on LLaVA-style pipelines connecting visual encoders to
language models \citep{videochatgpt2024,videollava2024}; recent families such
as InternVL, Qwen-VL, GLM-V, and Qwen3.5 scale this recipe with stronger dense
models, sparse MoE variants, dynamic resolution, early-fusion multimodal
training, and reasoning-oriented modes
\citep{internvl2024,internvl35_2025,qwen3vl2025,glm45v_2025,qwen3.5}. Because
these families differ in architecture and decoding conventions, Video-HOCA
treats evaluation design as part of the benchmark and reports taxonomy
agreement, Task-IV judge validation, macro-F1 and aggregation checks, and
temporal/decoding controls alongside the model scores.

\begin{figure}[t]
    \centering
    \includegraphics[width=\linewidth]{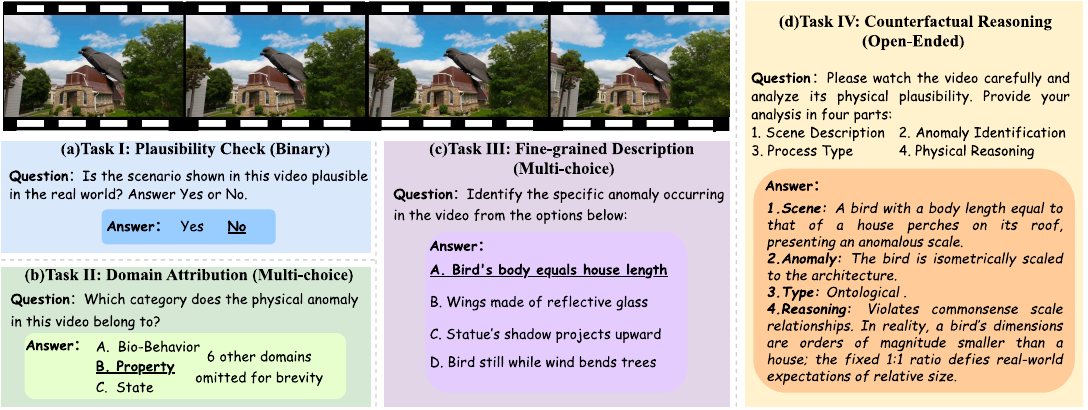}
    \caption{\textbf{Structured task design in Video-HOCA.} The benchmark
             evaluates physical anomaly reasoning through four progressive
             tasks: (Task~I) Plausibility Check, (Task~II) Anomaly Category
             Attribution, (Task~III) Fine-grained Anomaly Recognition, and
             (Task~IV) Open-ended Physical Reasoning. The sequence probes
             model capability from coarse plausibility detection to
             fine-grained recognition and mechanism-level explanation.}
    \label{fig:tasks}
\end{figure}

\section{Video-HOCA: Benchmark Design}

\subsection{Operational Taxonomy}

Video-HOCA uses an operational taxonomy: each anomaly is labeled by its
\emph{primary violated unit}. \textbf{Ontological} anomalies are
\emph{entity-centered}: the object or agent is internally inconsistent,
unstable, or capable of something its own definition should not allow.
\textbf{Causal} anomalies are \emph{relation-centered}: otherwise valid
entities interact with each other or with the environment in a way that violates
a physical regularity. This distinction is used purely as an operational
annotation rule for labeling the dataset.
Figure~\ref{fig:teaser} summarizes the two L1 branches and their L2 coverage.

The taxonomy has three levels. \textbf{L1} is the Ontological--Causal split used
for the main analyses. \textbf{L2} contains 9 coarse anomaly groups used for
balancing and annotation. \textbf{L3} contains the fine-grained anomaly tags
used to write options, reference answers, and metadata for individual videos.

\paragraph{L2 groups.}
The \textbf{Ontological} branch contains \emph{Property}, \emph{State}, and
\emph{Bio-Behavior}. The \textbf{Causal} branch contains \emph{Mechanics},
\emph{Material and Rigidity}, \emph{Mass and Volume Consistency},
\emph{Thermodynamics}, \emph{Chemistry}, and \emph{Optics}. These groups are
annotation bins rather than an exhaustive physical ontology; they are used to
balance the dataset and construct the tasks. Full definitions and canonical
examples are given in Appendix~\ref{app:taxonomy-details}; the resulting
taxonomy distribution is visualized in Figure~\ref{fig:taxonomy}. We test the
taxonomy by re-annotation: independent annotators must reproduce the labels
from the written rule alone.

\paragraph{Annotation agreement.}
\label{sec:taxonomy-agreement}

After freezing the guideline, three annotators re-labeled a stratified held-out
subset of 120 anomalous videos across the 9 L2 subcategories. Fleiss' $\kappa$
for the L1 Ontological-vs.-Causal split is $0.71$, indicating substantial
agreement; agreement on the 9-way L2 taxonomy is moderate, reflecting the
inherent subjectivity of fine-grained category boundaries. The L1 rule is
therefore reproducible, while the L2 task is more naturally read as a
diagnostic axis (Task~II) than as a sole capability claim.

\subsection{Data Construction and Verification}
\label{sec:dataset}

\textbf{Text-source candidates.} Figure~\ref{fig:pipeline} shows the construction
pipeline. Video-HOCA combines two anomaly-candidate streams. The first stream
starts from generated-video benchmarks that already provide clips with text
descriptions or human annotations of physical generation failures. We treat
these descriptions as candidates, map them into the Video-HOCA taxonomy with
GLM-4.5-Air \citep{glm45_2025}, apply a JSON schema check, and send the result
to human review.

\textbf{Video-source candidates.} The second stream starts from raw clips without
reliable anomaly labels. We generate part of it ourselves by asking
GPT-OSS-120B \citep{gptoss2025} to write counterfactual physical prompts and
using Wan~2.1-T2V-14B \citep{wan2025} to generate videos. We also reuse
unannotated generated videos from sources such as VBench-2.0
\citep{vbench2_2025}. Directly prompting video generators for targeted physical
anomalies is usually more efficient than searching through generic
generated-video datasets, but both routes still produce only candidates, not
ground truth.

\textbf{Coarse-to-fine verification.} For generated clips, a VLM first produces
frame-wise captions. An LLM then aggregates the temporal captions into an
anomaly summary, maps the summary to the operational taxonomy, and checks the
result with a JSON schema before human review. We found that direct video-level
VLM annotation often misses the anomalous event. On a 10-video pilot, direct
video-level annotation surfaced an anomaly in only 1 of 10 clips and even there
recovered the violation only partially, whereas the frame-wise captioning plus
temporal-aggregation pipeline identified the anomaly in 9 of 10 clips. This
higher recall comes with a precision cost: the pipeline also surfaces some
spurious or partially hallucinated anomaly candidates that are not actually
present in the video. We accept this trade-off because rejecting or
correcting model-suggested candidates during human review is substantially
cheaper than annotating every clip from scratch without any candidate, and
final labels and reference answers are decided by human reviewers using a
dedicated review tool (Figure~\ref{fig:anno-tool}) rather than by the
model pipeline.

\begin{figure}[t]
    \centering
    \includegraphics[width=\linewidth]{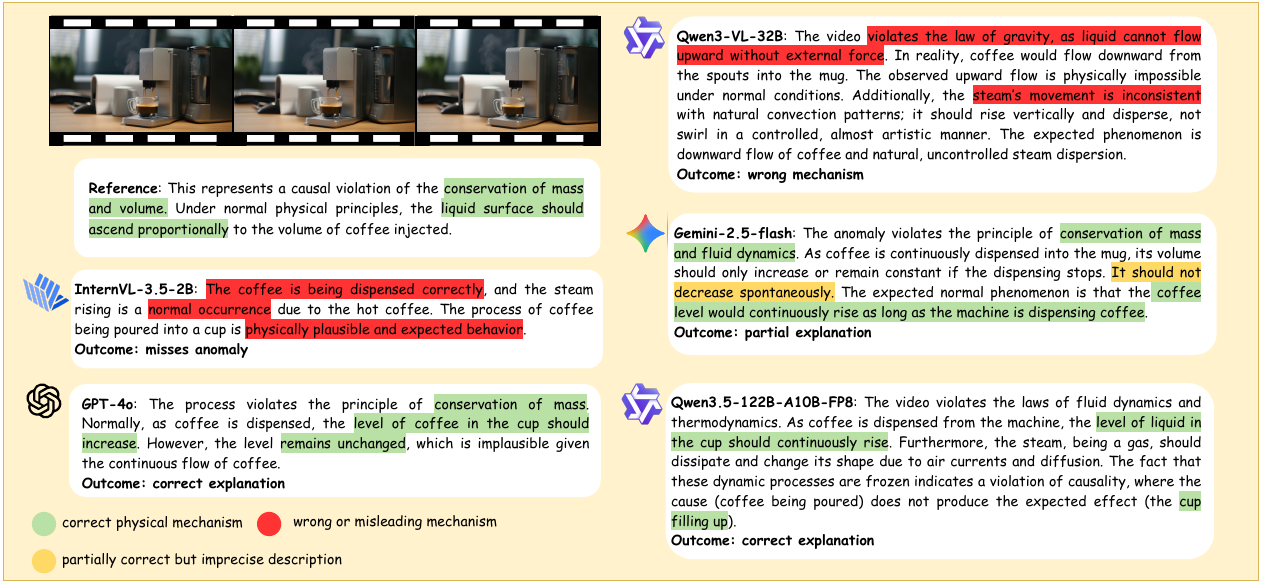}
    \caption{\textbf{Task-IV failure modes on a coffee-level anomaly.} All
             models observe the same video, where coffee is continuously
             dispensed but the liquid level does not rise. Green marks the
             correct conservation-based mechanism, red marks wrong or
             misleading mechanisms, and yellow marks partially correct but
             imprecise event descriptions. The example shows that models may
             recognize visible activity while missing or misattributing the
             physical mechanism.}
             \label{fig:qualitative}
    \vspace{-0.6em}
\end{figure}

\textbf{Dataset scale.} The anomaly-focused subset contains 809 videos: 630
single-anomaly clips, 59 temporal-grounding clips, and 120 multi-anomaly clips.
The 630 single-anomaly clips are balanced across the L3 tags, with 31--37
samples per fine-grained category. The 809 anomaly videos come from 15
generative sources; the largest contributors are Wan~2.1 (196 videos),
HunyuanVideo (129), Hailuo (123), and Sora (72). The source distribution is
shown in Figure~\ref{fig:source-dist}.

\textbf{Real-video matching.} We curate about 630 real-world clips from
Panda-70M \citep{panda70m} for the plausibility task, applying shot
segmentation, scene-change checks, resolution and caption filtering, and
realism checks; CLIP-based semantic matching \citep{clip2021} then pairs real
and anomalous videos by scene content so that Task~I is not solved by topic
alone. The final benchmark contains more than 1{,}400 videos and 3{,}470 QA
pairs, and the release package documents source type, taxonomy labels, task
tuples, splits, evaluation scripts, and the Task-IV judge protocol.

\paragraph{Diversity and coverage.}
Source diversity spans 15 generators (Wan~2.1, HunyuanVideo, Hailuo, Sora,
CogVideoX, Open-Sora, LTX-Video, Luma Dream Machine, Kling, Pika, Genmo,
Veo, VideoCrafter2, Cosmos, and others;
\citep{wan2025,hunyuanvideo2024,hailuo2024,sora2_2025,cogvideox2024,opensora2024,ltxvideo2024,lumadream2024,kling2024,pika2023,genmo2024,veo2024,videocrafter2_2024,cosmos2025})
plus real-world clips semantically matched to the anomaly set. Scene
diversity covers daily-life physical scenes with about 350 generation themes
and 200 unique scene types; anomaly diversity follows the Ontological--Causal
split and 9 L2 subcategories with roughly balanced sizes; task diversity
covers the four main QA tasks, temporal grounding, and multi-anomaly
reasoning. The taxonomy itself is validated by the re-annotation study in
\S\ref{sec:taxonomy-agreement}.

\subsection{Tasks and Evaluation}

\paragraph{Four tasks of increasing specificity.}
\textbf{Task I: Plausibility Check.} This binary task tests the coarsest level
of physical perception. Given a video, the model decides whether the observed
event is physically plausible. Real videos are semantically matched to anomalous
videos so that the task is not solved by scene topic alone.

\textbf{Task II: Anomaly Category Attribution.} Once a video is known to contain
an anomaly, the model maps it to one of the 9 L2 taxonomy groups. The prompt
includes short category definitions, and the model chooses among all L2 groups.
This task tests taxonomy attribution: the model may notice that something is
wrong, but still assign the wrong physical category.

\textbf{Task III: Fine-grained Anomaly Recognition.} This multi-choice task
pairs the ground-truth description with adversarial distractors that
GPT-OSS-120B generates from the original annotation and from other
taxonomy categories: each distractor is semantically close but physically
incorrect. The format is reliably solvable once the anomaly is identified
(human accuracy is $96.0$ in Table~\ref{tab:main}), so it isolates
fine-grained recognition from plausibility detection.

\textbf{Task IV: Open-ended Physical Reasoning.} The model gives a structured
answer with four parts: scene description, anomaly identification, anomaly type,
and physical explanation. The goal is to test whether the model can explain the
violated physical relation and the expected normal outcome, not only point to
the abnormal visual event.

Beyond the four main tasks, Video-HOCA includes two additional probes:
\emph{temporal grounding}, which asks for the anomalous time interval, and
\emph{multi-anomaly detection}, which asks models to identify concurrent
violations in the same video. We use these probes for auxiliary analysis rather
than for the main leaderboard. Temporal grounding contributes 200 interval-prediction
QA pairs, drawn from 59 newly annotated clips and 141 single-anomaly clips
that are reused from the main set with added temporal-interval labels (an
annotation example is shown in Figure~\ref{fig:temporal-anno}). Multi-anomaly
detection uses the 120 multi-anomaly clips with 64-frame input. We also run a
multi-frame scaling probe that compares the four main tasks under 16-frame and
64-frame inputs.

\paragraph{Task-IV judge protocol.}

Task~IV uses rubric-based evaluation against structured reference answers
rather than free-form commonsense judging. GPT-OSS-120B \citep{gptoss2025}
scores each response using fixed criteria---scene understanding, anomaly
identification, anomaly-type alignment, and physical-reasoning
quality---against the paired reference explanation. Full prompt text and
scoring templates are given in Appendix~\ref{app:judge}, and judge validation
(human--judge and judge--judge) is reported in \S\ref{sec:alt-metrics} and in
the main paper.

\paragraph{Metrics.}
Task~I reports binary-plausibility ACC and F1 on the balanced real/anomalous
split. Task~II and Task~III report macro-F1 on the Ontological and Causal
subsets: macro-F1 weights each taxonomy category equally and surfaces
systematic confusions that accuracy can hide even at roughly balanced counts
(\S\ref{sec:alt-metrics}). Task~IV uses the rubric-based judge score. For
compact reporting, we also compute a summary Overall column from Task~I F1,
Task~II/III macro-F1, and the Task~IV judge score, normalized via z-scores;
the summary is for readability only, and per-task results plus alternative
aggregations are reported in Table~\ref{tab:main} and \S\ref{sec:aggregation}.

\section{Experiments}

\subsection{Experimental Setup}

The main table uses Instruct-mode models only, which keeps the leaderboard
readable by comparing the default chat/instruct setting for each model rather
than mixing direct-answer and reasoning-trace models with different sampling
and token budgets. We evaluate 20 zero-shot Instruct configurations across
four families:
\emph{Qwen-VL/Qwen3.5} (Qwen2.5-VL-7B, Qwen3-VL-2B/8B/32B and Qwen3-VL-30B-A3B
\citep{qwen3vl2025}, Qwen3.5-9B/27B/35B-A3B/122B-A10B-FP8~\citep{qwen3.5});
\emph{InternVL} (InternVL-2.5-8B, InternVL-3.5-2B/8B/14B/38B and
InternVL-3.5-30B-A3B \citep{internvl2024,internvl35_2025});
\emph{GLM-4.6V} (GLM-4.6V-9B-Flash, GLM-4.6V-106B-FP8 \citep{glm45v_2025});
and \emph{closed-source} systems (GPT-4o \citep{gpt4o2024},
Gemini-2.5-Flash \citep{gemini25_2025}, Gemini-3-Flash~\citep{gemini3flash_2025}). Task~IV answers
are scored against structured reference explanations by GPT-OSS-120B
\citep{gptoss2025}; we also report a
DeepSeek-V4-Flash~\citep{deepseekai2026deepseekv4} judge variant, denoted as
Task~IV (D), when available.

\paragraph{Decoding and video input.}
The Instruct-mode main table uses deterministic decoding (temperature $=0$,
no sampling, $1{,}024$-token budget) on all four tasks; Thinking-mode results
are kept out of the leaderboard. The Thinking-vs.-Instruct controls in
\S\ref{sec:thinking-vs-instruct} use vendor-recommended sampling and an
$8{,}192$-token budget (full settings in Appendix~\ref{app:details}). Unless
otherwise noted, model inputs use 16 uniformly sampled frames; the
multi-frame scaling probe additionally uses 64 frames, and multi-anomaly
detection uses 64-frame input. Single-frame and shuffled-frame variants
(\S\ref{sec:input-controls}) are temporal-confound controls, not part of
the leaderboard.

\subsection{Main Results}

Table~\ref{tab:main} gives the four-task Instruct results. Under z-score
aggregation, Qwen3.5-27B (61.2) and Gemini-3-Flash (60.2) lead the Overall
column, with Qwen3.5-122B-A10B-FP8 and Qwen3-VL-32B at 57.6/57.4; positions
3--5 swap under raw weighted aggregation (Appendix~\ref{app:aggregation}).
Scale or recency does not remove the need for task-level analysis. Task~II
is the clearest attribution bottleneck: across the 20 configurations,
macro-F1 ranges from $14.7$ to $50.2$ on Ontological and from $1.3$ to
$65.2$ on Causal, with cross-model means of $36.5/44.3$---well below the
Task~III means of $70.1/63.2$. Task~IV exposes a different failure mode:
models often identify visible activity but miss or misattribute the physical
mechanism, as shown qualitatively in Figure~\ref{fig:qualitative}. Claims in
this paper come from the task-level results and the aggregation checks in
\S\ref{sec:aggregation}.

\begin{table}[t]
    \centering
    \caption{Main Instruct-mode evaluation results on Video-HOCA. Task~I
    reports binary-plausibility ACC/F1 (\%); Task~II and Task~III report macro-F1
    (\%) on Ontological--Causal subsets; Task~IV reports
    rubric-based judge scores (0--100) decomposed into
    Ontological--Causal dimensions. Task~IV (D) uses DeepSeek-V4-Flash as
    the judge. ``Overall'' first standardizes each main task score with
    z-scores computed over the main-table models, averages the four
    standardized scores, and linearly rescales the result as $50+10z$ for
    readability. Bold indicates the best model result in each column, and
    underlining indicates the second best; the Human row is excluded from this
    marking.
    All rows use deterministic Instruct-mode decoding; Thinking-mode
    results are analyzed separately in
             \S\ref{sec:thinking-vs-instruct}.} \label{tab:main} \scriptsize
             \setlength{\tabcolsep}{2.4pt}
    \resizebox{\linewidth}{!}{%
    \begin{tabular}{>{\centering\arraybackslash}p{2.9cm}cccccccccc>{\centering\arraybackslash}p{0.9cm}}
        \toprule
        \multirow{2}{*}{\textbf{Model}} & \multicolumn{2}{c}{\textbf{Task~I}} & \multicolumn{2}{c}{\textbf{Task~II}} & \multicolumn{2}{c}{\textbf{Task~III}} & \multicolumn{2}{c}{\textbf{Task~IV}} & \multicolumn{2}{c}{\textbf{Task~IV (D)}} & \multirow{2}{*}{\textbf{Overall}} \\
        \cmidrule(lr){2-3}\cmidrule(lr){4-5}\cmidrule(lr){6-7}\cmidrule(lr){8-9}\cmidrule(lr){10-11}
                                      & \textbf{ACC} & \textbf{F1} & \textbf{Onto.} & \textbf{Caus.} & \textbf{Onto.} & \textbf{Caus.} & \textbf{Onto.} & \textbf{Caus.} & \textbf{Onto.} & \textbf{Caus.} & \\
        \midrule
        Human                & --     & --                          & --                          & --                           & \multicolumn{2}{c}{96.0}    & \multicolumn{2}{c}{81.4} & \multicolumn{2}{c}{84.9} & --                            \\
        \midrule
        \rowcolor{black!8}\multicolumn{12}{c}{\emph{Open-weight dense models ($<$7B)}} \\
        InternVL-3.5-2B         & 61.7 & 69.6 & 25.0 & 49.6 & 70.4 & 55.6 & 40.9 & 47.1 & 39.7 & 44.6 & 43.3 \\
        Qwen3-VL-2B             & 71.7 & 75.0 & 34.9 & 46.6 & 73.1 & 67.8 & 50.2 & 51.0 & 44.6 & 44.6 & 48.0 \\
        \midrule
        \rowcolor{black!8}\multicolumn{12}{c}{\emph{Open-weight dense models ($\geq$7B)}} \\
        InternVL-2.5-8B         & 56.1 & 62.9 & 14.7 & 5.4 & 15.1 & 11.1 & 1.9 & 1.7 & 1.3 & 1.7 & 21.6 \\
        Qwen2.5-VL-7B           & 71.4 & 76.5 & 34.3 & 53.1 & 67.9 & 60.9 & 52.2 & 54.3 & 48.1 & 49.2 & 49.0 \\
        InternVL-3.5-8B         & 67.8 & 74.7 & 40.5 & 49.4 & 74.2 & 67.4 & 53.7 & 47.2 & 46.0 & 41.2 & 48.6 \\
        Qwen3-VL-8B             & 75.4 & 78.2 & 39.8 & 49.7 & 83.7 & 76.6 & 64.6 & 63.7 & 62.9 & 59.1 & 53.5 \\
        Qwen3.5-9B              & 79.8 & 82.6 & 34.2 & 30.9 & 74.9 & 67.1 & 75.3 & 58.3 & 69.7 & 51.7 & 51.9 \\
        InternVL-3.5-14B        & 73.0 & 77.6 & 40.7 & 59.8 & 77.7 & 69.9 & 63.2 & 55.7 & 57.0 & 54.4 & 52.8 \\
        Qwen3.5-27B             & \underline{85.6} & \underline{86.2} & \underline{47.7} & \textbf{65.2} & \textbf{90.2} & \textbf{85.7} & \underline{77.5} & \underline{65.1} & 75.7 & 62.2 & \textbf{61.2} \\
        Qwen3-VL-32B            & 79.5 & 82.3 & 46.0 & 55.6 & 84.1 & 79.0 & \textbf{78.5} & 62.9 & \underline{75.9} & 59.3 & 57.4 \\
        InternVL-3.5-38B        & 75.6 & 77.5 & 31.8 & 54.6 & 78.3 & 77.4 & 67.8 & 54.8 & 60.4 & 53.7 & 52.5 \\
        \midrule
        \rowcolor{black!8}\multicolumn{12}{c}{\emph{Open-weight MoE and 100B-class FP8 deployments}} \\
        GLM-4.6V-9B-Flash       & 63.6 & 72.6 & 38.8 & 47.3 & 78.0 & 67.4 & 71.0 & 54.8 & 68.4 & 52.3 & 49.4 \\
        InternVL-3.5-30B-A3B    & 68.7 & 74.4 & 43.8 & 51.2 & 82.1 & 71.2 & 62.7 & 48.3 & 54.8 & 46.6 & 50.3 \\
        Qwen3-VL-30B-A3B        & 78.6 & 80.8 & 46.4 & 50.2 & 81.6 & 73.4 & 74.6 & 64.4 & 69.8 & 58.8 & 55.6 \\
        Qwen3.5-35B-A3B         & 82.9 & 84.6 & 21.9 & 1.3 & 32.7 & 25.3 & 75.8 & 61.1 & 74.9 & 58.2 & 43.8 \\
        GLM-4.6V-106B-FP8       & 69.8 & 76.2 & 31.5 & 21.9 & 24.9 & 18.0 & 71.0 & 58.9 & 70.5 & 57.3 & 41.8 \\
        Qwen3.5-122B-A10B-FP8   & \textbf{88.0} & \textbf{87.5} & 45.0 & 41.0 & \underline{85.9} & 79.2 & 73.1 & 61.4 & 69.0 & 58.8 & 57.6 \\
        \midrule
        \rowcolor{black!8}\multicolumn{12}{c}{\emph{Closed-source systems}} \\
        GPT-4o                  & 74.0 & 75.2 & 19.5 & 33.1 & 59.3 & 57.1 & 69.4 & 63.0 & 71.0 & 63.9 & 46.5 \\
        Gemini-2.5-Flash        & 75.5 & 77.7 & 42.8 & 59.9 & 82.8 & 73.0 & 73.6 & \underline{65.6} & 75.6 & \underline{65.1} & 55.1 \\
        Gemini-3-Flash          & 84.4 & 85.5 & \textbf{50.2} & \underline{60.9} & 85.6 & \underline{80.9} & 76.6 & \textbf{67.6} & \textbf{76.4} & \textbf{66.8} & \underline{60.2} \\
        \bottomrule
    \end{tabular}}
\end{table}

\paragraph{Ontological--Causal differences.}
The Ontological--Causal gap is task-conditional. On Task~III and Task~IV,
strong Instruct models usually do better on Ontological subsets than on
Causal subsets (Qwen3.5-27B, for instance, scores $90.2/85.7$ on Task~III and
$77.5/65.1$ on Task~IV); the gap is larger on open-ended explanation, where
Causal violations require the model to link entities, forces,
environments, and expected outcomes. Task~II asks for a taxonomy label rather
than an explanation, and behaves differently: Causal cases often have visible
cues (collision, gravity, fluid, shadow), while Ontological cases spread
across shape, material, state, and biology. The cross-model mean is reversed
(Causal $44.3$ vs.\ Ontological $36.5$ macro-F1), but several MoE and
100B-class FP8 configurations (Qwen3.5-9B, Qwen3.5-35B-A3B,
Qwen3.5-122B-A10B-FP8, GLM-4.6V-106B-FP8) score higher on Ontological than
Causal in Task~II. We therefore no longer treat Causal anomalies as uniformly
harder; the direction depends on both the task and the model configuration.

\subsection{Evaluation Validation}

\paragraph{Human--judge and judge--judge validation.}
On 100 sampled Task-IV responses from four representative models, human
scoring vs.\ GPT-OSS-120B yields Pearson $=0.8957$ and Spearman $=0.8914$;
human-authored answers receive 81.4 under GPT-OSS-120B and 84.9 under
DeepSeek-V4-Flash, supporting the rubric protocol from both the human and
judge-substitution sides.

\paragraph{Alternative metrics.}
\label{sec:alt-metrics}
We use macro-F1 for Task~II and Task~III because accuracy can hide uneven
category behavior. The distinction matters most for Task~II: cross-model means
change from $34.0/48.8$ accuracy to $36.5/44.3$ macro-F1 on
Ontological--Causal subsets, showing that Causal attribution looks stronger
under accuracy than under category-balanced scoring. Task~III is less
sensitive: the means change from $73.9/67.7$ accuracy to $70.1/63.2$
macro-F1, and the recognition-over-attribution gap remains. The released metric
code computes both accuracy and macro-F1 from the same predictions.

\paragraph{Aggregation sensitivity.}
\label{sec:aggregation}
The summary score is used only for readability. As shown in
Appendix~\ref{app:aggregation}, switching the Task-IV judge from GPT-OSS to
DeepSeek leaves the z-score ranking almost unchanged, and replacing z-score
aggregation with a raw weighted average produces only local swaps among
mid-ranked models. The top two models stay the same under all four scoring
variants.

\subsection{Confound Analyses}
If Causal anomalies were harder only because they are more dynamic, the
benchmark would mostly be measuring temporal integration. We test this
directly.

\paragraph{Static/dynamic distribution.}
We manually stratified the 630 single-anomaly videos by branch and temporal
demand: a \emph{dynamic} item requires cross-frame motion cues, while a
\emph{static} item can be judged from a representative frame.
Table~\ref{tab:static-dynamic-summary} reports the split and Task~III/IV means
over the main model set. Causal anomalies are more dynamic on average, but
Ontological anomalies are not simply the static half of the dataset.

\paragraph{Static causal vs.\ ontological.}
\label{sec:static-vs-onto}
Restricting to static-only videos removes the temporal advantage of
Ontological cases. The gap shrinks substantially but does not disappear: on
Task~IV, the cross-model Ontological--Causal gap drops from about $8.4$ points
on the full set to $3.6$ points on the static-only subset
(Table~\ref{tab:static-dynamic-summary}). A sizeable fraction of the gap is
therefore attributable to temporal demand, but static Causal cases still trail
static Ontological cases on reasoning-heavy tasks. Per-model results are
in Appendix~\ref{app:static-dynamic-full}.

\begin{table}[t]
    \centering
    \caption{Static/dynamic analysis by anomaly branch. Scores are averaged
             over the main model set. Static Causal cases remain below Static
             Ontological cases, so the Ontological--Causal gap is not reducible
             to temporal demand alone.}
    \label{tab:static-dynamic-summary}
    \footnotesize
    \setlength{\tabcolsep}{7pt}
    \begin{tabular}{llccc}
        \toprule
        Branch & Temporal type & Videos & Task~III & Task~IV \\
        \midrule
        Ontological & Static  & 130 & 71.2 & 63.9 \\
                    & Dynamic & 124 & 68.7 & 63.6 \\
        Causal      & Static  & 128 & 69.4 & 60.3 \\
                    & Dynamic & 248 & 59.9 & 53.0 \\
        \midrule
        Static gap (Onto. -- Caus.) & -- & -- & +1.8 & +3.6 \\
        \bottomrule
    \end{tabular}
\end{table}

\paragraph{Input controls.}
\label{sec:input-controls}
We also vary temporal input for Qwen3.5-122B-A10B-FP8: original 16-frame
order, a single middle frame, and 16 frames in shuffled order. The
middle-frame setting hurts most tasks while shuffling stays close to the
original order, suggesting that the model uses multi-frame coverage more
reliably than precise frame order
(Appendix~\ref{app:temporal-frame-ablation}).

\paragraph{Decoding controls for Thinking vs.\ Instruct.}
\label{sec:thinking-vs-instruct}
Thinking modes change both the generation process and the decoding budget:
they emit an explicit reasoning trace and usually require sampling with a
larger token budget, so greedy decoding can make these models loop or fail to
finish. To isolate reasoning mode from decoding, we compare three settings:
\emph{Instruct} (the deterministic setting in Table~\ref{tab:main}),
\emph{Instruct-Align} (Instruct rerun with the corresponding Thinking
sampling parameters and output budget), and \emph{Thinking} (the
vendor-recommended native setting). Table~\ref{tab:thinking-align-instruct}
asks whether native Thinking gains can be reproduced by making Instruct
decoding more thinking-like.

\begin{table}[t]
    \centering
    \caption{Comparison of Instruct, Instruct-Align, and native Thinking decoding on Video-HOCA. Task~I reports ACC/F1 (\%); Task~II and Task~III report Ontological--Causal macro-F1 (\%); Task~IV reports rubric-based Ontological--Causal judge scores (0--100). Diag. Overall uses the same fixed main-table z-score normalization as Table~\ref{tab:main}; bold marks the best setting within each model block.}
    \label{tab:thinking-align-instruct}
    \scriptsize
    \setlength{\tabcolsep}{3.0pt}
    \resizebox{\linewidth}{!}{%
    \begin{tabular}{llccccccccc}
        \toprule
        \multirow{2}{*}{\textbf{Model}} & \multirow{2}{*}{\textbf{Setting}} & \multicolumn{2}{c}{\textbf{Task~I}} & \multicolumn{2}{c}{\textbf{Task~II}} & \multicolumn{2}{c}{\textbf{Task~III}} & \multicolumn{2}{c}{\textbf{Task~IV}} & \multirow{2}{*}{\textbf{Diag. Overall}} \\
        \cmidrule(lr){3-4}\cmidrule(lr){5-6}\cmidrule(lr){7-8}\cmidrule(lr){9-10}
          & & \textbf{ACC} & \textbf{F1} & \textbf{Onto.} & \textbf{Caus.} & \textbf{Onto.} & \textbf{Caus.} & \textbf{Onto.} & \textbf{Caus.} & \\
        \midrule
        GLM-4.6V-106B-FP8            & Instruct       & \textbf{69.8} & \textbf{76.2} & 31.5 & 21.9 & 24.9 & 18.0 & 71.0 & 58.9 & 41.8 \\
        GLM-4.6V-106B-FP8            & Instruct-Align & 68.4 & 75.1 & 36.2 & 36.9 & 32.5 & 26.8 & 73.6 & 60.7 & 44.6 \\
        GLM-4.6V-106B-FP8            & Thinking       & 68.5 & 75.7 & \textbf{45.7} & \textbf{59.8} & \textbf{78.9} & \textbf{72.6} & \textbf{78.3} & \textbf{62.0} & \textbf{54.2} \\
        \midrule
        Qwen3.5-122B-A10B-FP8        & Instruct       & \textbf{88.0} & 87.5 & 45.0 & 41.0 & 85.9 & 79.2 & 73.1 & 61.4 & 57.6 \\
        Qwen3.5-122B-A10B-FP8        & Instruct-Align & 86.7 & \textbf{87.7} & 44.6 & 35.4 & \textbf{88.6} & 80.3 & 68.0 & 61.5 & 56.9 \\
        Qwen3.5-122B-A10B-FP8        & Thinking       & 83.7 & 86.0 & \textbf{52.4} & \textbf{50.9} & 86.6 & \textbf{82.6} & \textbf{78.6} & \textbf{62.4} & \textbf{59.3} \\
        \midrule
        Qwen3-VL-32B                 & Instruct       & \textbf{79.5} & \textbf{82.3} & 46.0 & \textbf{55.6} & 84.1 & 79.0 & 78.5 & \textbf{62.9} & \textbf{57.4} \\
        Qwen3-VL-32B                 & Instruct-Align & 78.7 & 81.5 & 44.8 & 52.6 & 84.4 & 76.7 & \textbf{80.5} & 61.9 & 56.5 \\
        Qwen3-VL-32B                 & Thinking       & 74.8 & 79.4 & \textbf{50.9} & 58.8 & \textbf{86.9} & \textbf{79.3} & 76.6 & 61.9 & 56.8 \\
        \bottomrule
    \end{tabular}}
\end{table}

Native Thinking improves reasoning-heavy tasks for some models, but the
pattern is not reducible to output length or chain-of-thought style alone:
GLM-4.6V-106B-FP8 and Qwen3.5-122B-A10B-FP8 benefit from native Thinking on
several Task~II--IV columns, while their Instruct-Align rows do not
consistently reproduce those gains, and Qwen3-VL-32B's Instruct-Align row
stays close to Instruct without dominating native Thinking. We therefore
treat Thinking as a diagnostic comparison of decoding and model mode, not as
a leaderboard axis.

\section{Limitations and Release}

Video-HOCA is diagnostic, and its scores should be read with that scope in
mind. Synthetic artifacts remain a concern: many anomalous videos come from
current video generators, so some model differences may reflect generator
artifacts rather than anomaly reasoning. The taxonomy has subjective edges
at the fine-grained category boundaries, where annotators can reasonably
disagree on which L2 group best captures a given anomaly; we therefore
treat the Task~II attribution score as one diagnostic axis among four
rather than as a sole capability claim. Task~IV depends on an LLM judge,
which we validate against human scoring and a substitute judge but cannot
eliminate fully. An anonymized artifact package accompanies the submission;
release contents and asset restrictions are in
Appendix~\ref{app:artifact-release}.

\section{Conclusion}

Video-HOCA is a diagnostic benchmark for physical anomaly reasoning, with a
validated taxonomy and evaluation pipeline. Across 20 Instruct-mode Video-LLM
configurations, we find three bounded patterns. First, models can detect
anomalies before they can attribute or explain them: Task~I scores cluster
around $75$--$88$ while Task~II macro-F1 stays mostly below $50$. Second, the
Ontological--Causal gap depends on the task and on the model configuration
rather than pointing in one direction everywhere. Third, temporal demand,
frame ordering, and Thinking-style decoding all change how the results should
be interpreted. These claims are intentionally bounded: Video-HOCA is meant
to make claims about physical anomaly reasoning testable, and to demarcate
where those claims stop.

\bibliographystyle{plainnat}
\bibliography{hoca_refs}

\clearpage
\appendix

\section{Appendix}

\subsection{Annotation Interface and Dataset Visualizations}
\label{app:additional-figures}

\begin{figure}[t]
    \centering
    \includegraphics[width=\linewidth]{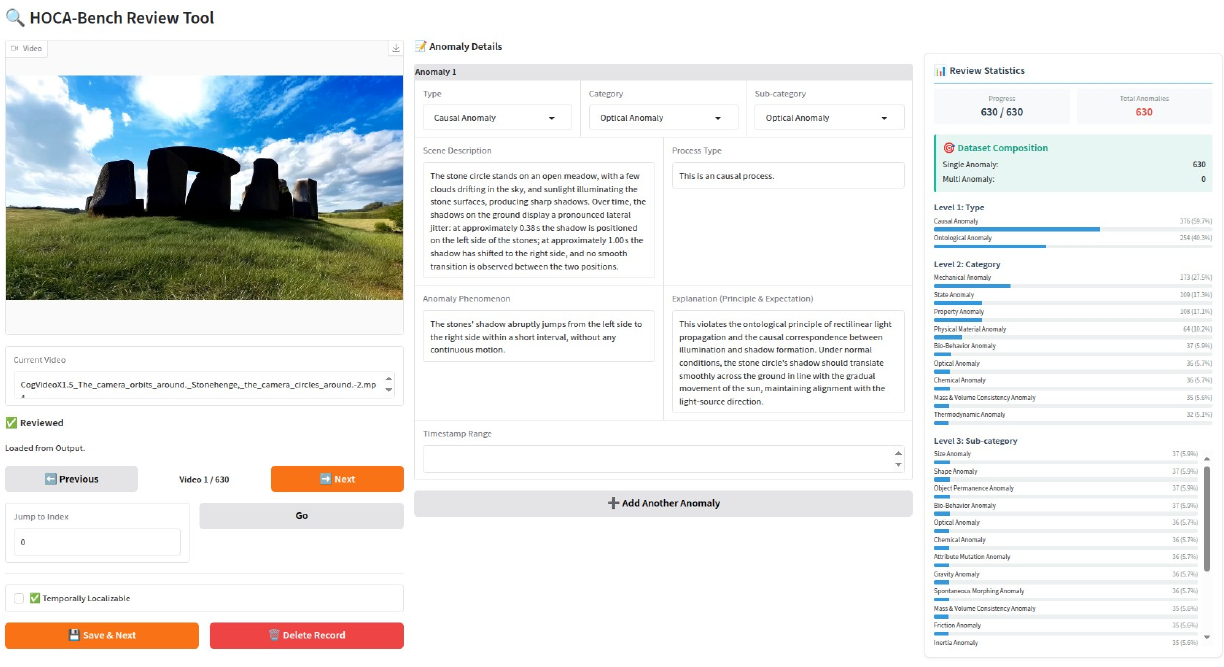}
    \caption{\textbf{Video-HOCA annotation interface.} The browser-based
             review tool used by annotators. It pairs a video player with
             editable fields for taxonomy type, category, sub-category,
             scene description, anomaly phenomenon, principle/expectation,
             and an optional temporal-localization range. The right panel
             tracks per-category dataset progress.}
    \label{fig:anno-tool}
\end{figure}

\begin{figure}[t]
    \centering
    \includegraphics[width=0.92\linewidth]{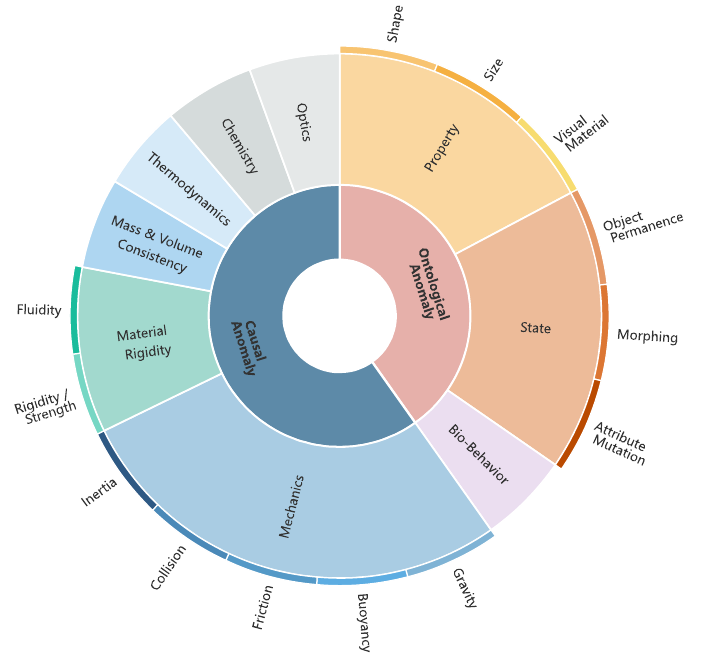}
    \caption{\textbf{Taxonomy distribution.} The sunburst chart visualizes
             the Ontological--Causal branches and their fine-grained L2
             categories.} \label{fig:taxonomy}
\end{figure}

\begin{figure}[t]
    \centering
    \includegraphics[width=0.82\linewidth]{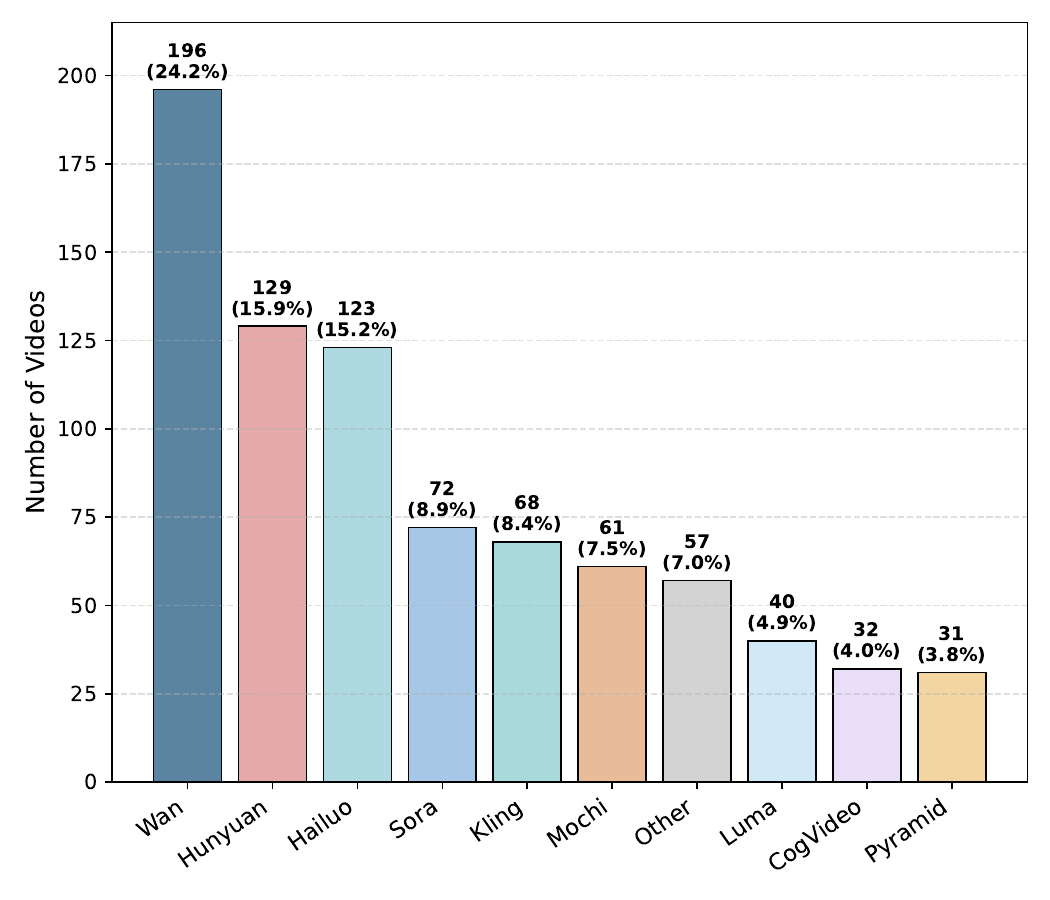}
    \caption{\textbf{Video-source distribution.} The generative sources used
             for Video-HOCA anomaly clips, supporting the source-diversity
             statistics reported in the main text.}
    \label{fig:source-dist}
\end{figure}

\begin{figure}[t]
    \centering
    \includegraphics[width=0.86\linewidth]{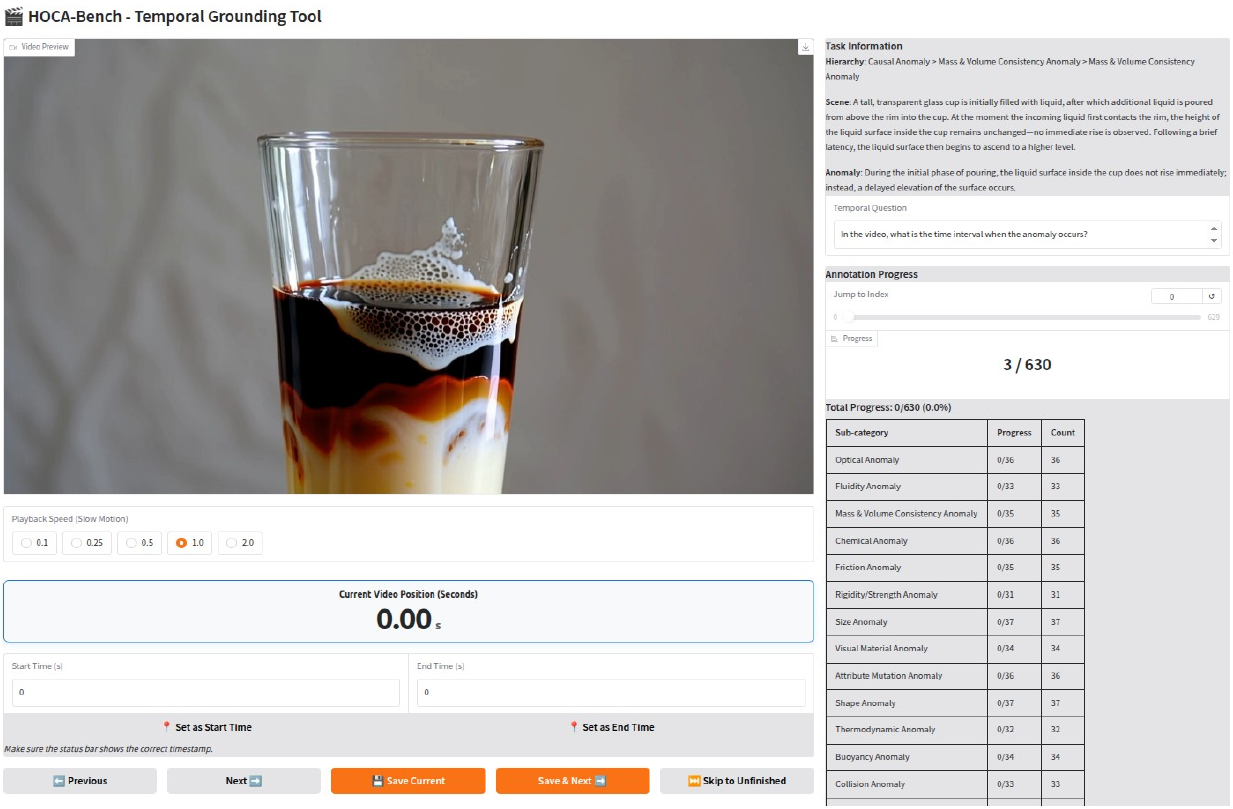}
    \caption{\textbf{Temporal annotation example.} Annotated time intervals
             for the temporal-grounding probe. The probe localizes when
             anomalous events occur and is auxiliary to the main
             Ontological--Causal claims.}
    \label{fig:temporal-anno}
\end{figure}

This subsection collects the supporting visualizations referenced in the main
paper: the human review tool used by annotators
(Figure~\ref{fig:anno-tool}), the taxonomy distribution
(Figure~\ref{fig:taxonomy}), the generative-source distribution
(Figure~\ref{fig:source-dist}), and an example of the temporal-grounding
annotation (Figure~\ref{fig:temporal-anno}).

\subsection{Detailed Taxonomy Guidelines}
\label{app:taxonomy-details}

Our annotation pipeline follows the same operational principle used in the main
paper: the final label is determined by the \emph{primary violated unit}, not
by the most visually salient event alone. Annotators first verify that a video
contains a physically implausible event, then summarize the anomalous event in
natural language, assign the anomaly to the Ontological or Causal branch using
the rules in \S\ref{sec:taxonomy-agreement}, and finally map it to one of nine
L2 subcategories. Samples without a stable primary violated unit are marked for
adjudication and are not used as canonical examples.

For the L1 Ontological-vs.-Causal decision, the practical rule is:
\begin{itemize}[leftmargin=1.5em]
    \item assign \textbf{Ontological} if the dominant violation concerns an entity's
          own identity, intrinsic property, state stability, or biological capability;
    \item assign \textbf{Causal} if the dominant violation concerns an interaction law,
          contact relation, force law, or environment-mediated transition between
          otherwise valid entities.
\end{itemize}

\begin{table}[t]
    \centering
    \caption{Full Video-HOCA taxonomy. L1 is the Ontological--Causal split;
             L2 is the 9 anomaly groups; L3 lists the fine-grained tags
             actually used as labels. For the four L2 groups with no further
             subdivision (Bio-Behavior, Mass and Volume Consistency,
             Thermodynamics, Chemistry, Optics), the L3 tag is identical to
             the L2 name. Canonical examples are selected to have a stable
             primary violated unit.}
    \label{tab:taxonomy-def}
    \footnotesize
    \setlength{\tabcolsep}{4pt}
    \renewcommand{\arraystretch}{1.15}
    \begin{tabularx}{\linewidth}{>{\raggedright\arraybackslash}p{2.4cm}
                                  >{\raggedright\arraybackslash}p{2.6cm}
                                  >{\raggedright\arraybackslash}X
                                  >{\raggedright\arraybackslash}X}
        \toprule
        \textbf{L2 group} & \textbf{L3 tag} & \textbf{Primary violation} & \textbf{Canonical example} \\
        \midrule
        \rowcolor{black!8}
        \multicolumn{4}{l}{\textbf{\emph{Ontological anomalies (entity-centered violations)}}} \\
        \multirow{3}{=}{\textbf{Property}}
           & Shape           & Geometrically impossible or contradictory shape.        & Three-headed goat. \\
           & Size            & Object scale violates physical bounds.                   & Tiny elephant on a dinner plate. \\
           & Visual Material & Surface appearance inconsistent with entity identity.    & Turtle with a ruby-like shell; transparent wood. \\
        \midrule
        \multirow{3}{=}{\textbf{State}}
           & Object Permanence  & Spontaneous appearance or disappearance of the whole entity. & Chair appears out of thin air. \\
           & Morphing           & Shape or size changes without external cause.                & Paper grows longer by itself. \\
           & Attribute Mutation & A visual attribute is temporally inconsistent.               & Butterfly wings change from yellow to white. \\
        \midrule
        \textbf{Bio-Behavior} & Bio-Behavior &
           A biological agent violates its own anatomy, motor ability, or species-level capability.
           & Human joints bend backward; eyeballs detach from sockets and swing. \\
        \midrule
        \rowcolor{black!8}
        \multicolumn{4}{l}{\textbf{\emph{Causal anomalies (relation-centered violations)}}} \\
        \multirow{5}{=}{\textbf{Mechanics}}
           & Gravity   & Universal gravitation fails (e.g., upward fall, no pull).         & Mug floats off the table; ball rolls uphill. \\
           & Buoyancy  & Archimedes' principle is violated.                                 & Heavy stone floats on water. \\
           & Friction  & Surface resistance is incorrect.                                   & Object slides indefinitely on grass. \\
           & Collision & Impenetrability or momentum transfer fails.                        & Solid objects pass through each other. \\
           & Inertia   & Newton's first law is violated.                                    & Object stops or reverses without external force. \\
        \midrule
        \multirow{2}{=}{\textbf{Material and Rigidity}}
           & Rigidity/Strength & A solid loses rigidity or deforms abnormally.            & Glass bends like rubber. \\
           & Fluidity          & Flow, viscosity, or surface tension behaves incorrectly. & Square water droplets; liquid behaves like a solid. \\
        \midrule
        \textbf{Mass and Volume Consistency} & Mass and Volume Consistency &
           Matter or volume changes during interaction without a plausible cause.
           & Water volume increases while pouring; cutting an object increases total volume. \\
        \midrule
        \textbf{Thermodynamics} & Thermodynamics &
           Temperature-dependent state changes fail or occur under the wrong condition.
           & Ice remains solid in boiling water; liquid boils without a heat source. \\
        \midrule
        \textbf{Chemistry} & Chemistry &
           Combustion or chemical transformation fails under ordinary reaction conditions.
           & Paper does not burn in flame; wood turns to ash without heat. \\
        \midrule
        \textbf{Optics} & Optics &
           Light transport, shadow, or reflection violates geometric constraints.
           & Shadow points toward the light source; mirror shows an impossible reflection. \\
        \bottomrule
    \end{tabularx}
\end{table}

\paragraph{Adjudication rule.} Canonical examples are selected only when the primary
violated unit is stable under the written guideline. If an item can be
reasonably explained by multiple rule layers, annotators do not force it into a
canonical category. They first identify the dominant violation; if no dominant
reading is stable, the item is flagged for adjudication or excluded from the
scored canonical set. This policy is meant to keep the benchmark focused on
physical anomaly reasoning rather than on ambiguous taxonomy edge cases.

\paragraph{Taxonomy attribution prompt.} The following prompt is used after a
video has been summarized into a short physical-realism analysis. It is given
both to the LLM annotation pipeline and to human annotators as the written
classification guideline. L2/L3 names are unified with
Table~\ref{tab:taxonomy-def}; the decision procedure is entity-centered
vs.\ relation-centered, matching the operational rule in
\S\ref{sec:taxonomy-agreement}.

{\small
\begin{verbatim}
You are an expert video anomaly analyst. Your task is to analyze a video
based on a provided "Analysis Summary" and classify the primary physical
anomaly according to the Video-HOCA taxonomy.

### 1. Taxonomy Structure
- Ontological anomaly (entity-centered): the dominant violation lies in
  an entity's own identity, intrinsic property, state, or biological
  capability, regardless of any interaction the entity participates in.
    * Property:     Shape, Size, Visual Material.
    * State:        Object Permanence, Morphing,
                    Attribute Mutation.
    * Bio-Behavior: anatomy / motor / species-level capability
                    violations of a biological agent.
- Causal anomaly (relation-centered): the dominant violation lies in a
  physical relation between otherwise valid entities, or between an
  entity and the environment.
    * Mechanics:                    Gravity, Buoyancy, Friction,
                                    Collision, Inertia.
    * Material and Rigidity:        Rigidity, Fluidity.
    * Mass and Volume Consistency:  matter / volume changes during
                                    interaction without a plausible cause.
    * Thermodynamics:               temperature-dependent state changes
                                    that fail or occur under the wrong
                                    condition.
    * Chemistry:                    combustion or chemical transformation
                                    that fails under ordinary reaction
                                    conditions.
    * Optics:                       light transport, shadow, or
                                    reflection violates geometric
                                    constraints.

### 2. Decision Procedure
Step 1. Decide L1 (Ontological vs Causal) by identifying the
        *primary violated unit*:
          - If the violation is entity-centered (intrinsic to one
            entity's identity, property, state, or biological
            capability), assign Ontological.
          - If the violation is relation-centered (between entities,
            or between an entity and the environment), assign Causal.
          - The presence of interaction does NOT automatically imply
            Causal. A clip can show two entities interacting while
            the dominant violation is still intrinsic to one of them
            (e.g., a person whose joints bend backward while walking:
            the contact is normal, the bio-behavior violation is
            intrinsic).
Step 2. Within the chosen L1 branch, assign the L2 group whose written
        definition the violation matches most precisely. If two groups
        apply equally well, mark the item for adjudication rather than
        forcing a label.
Step 3. Within L2, assign the most specific L3 sub-category from the
        list in Section 1.

### 3. Annotation Principles
1. Visual honesty: describe exactly what is shown, regardless of how
   impossible it is.
2. Setup vs observation: scene_description states the stage BEFORE the
   anomaly; anomaly_phenomenon describes the "wrong" event.
3. Dual-part reasoning ("Principle + Expectation" template) for
   normal_phenomenon. Examples:
   - Gravity: violates the Law of Universal Gravitation. Normally, an
     unsupported object undergoes vertical downward displacement.
   - Collision: violates impenetrability / momentum transfer. Normally,
     the subject should stop or rebound on contact rather than clip
     through.
   - Object Permanence: violates object constancy. Normally,
     macroscopic objects persist unless a specific physical process
     intervenes.
   - Rigidity: violates the property of solid rigidity. Normally, a
     solid resists deformation and keeps structural integrity under
     contact.

### 4. Few-Shot Examples
Example 1
Summary: A mug sits on a table, then suddenly flies up into the air.
Output:
{
  "type": "Causal",
  "l2_group": "Mechanics",
  "l3_tag": "Gravity",
  "scene_description": "A wooden table with a ceramic mug placed in
                        the center.",
  "anomaly_phenomenon": "The mug lifts off the table and floats
                         upwards without any external force.",
  "primary_violated_unit": "the gravitational relation between the
                            mug and the Earth",
  "normal_phenomenon": "The mug should remain stationary on the
                       table; once unsupported it should fall under
                       gravity, not rise."
}

Example 2
Summary: A person is walking, and their shirt suddenly changes from
blue to red.
Output:
{
  "type": "Ontological",
  "l2_group": "State",
  "l3_tag": "Attribute Mutation",
  "scene_description": "A person wearing a blue shirt is walking
                        across a room.",
  "anomaly_phenomenon": "The shirt's color suddenly changes from blue
                         to red without any lighting change.",
  "primary_violated_unit": "the temporal stability of the shirt's
                            color attribute",
  "normal_phenomenon": "An object's color should remain temporally
                       consistent unless acted upon by external light
                       or chemical changes."
}

### 5. Current Task
Analyze the following summary and produce the JSON described above.

Summary:
[INSERT SUMMARY TEXT HERE]

Output (JSON only, no extra text):
\end{verbatim}
}

\subsection{Task-III Distractor Generation Prompt}
\label{app:distractor}

Task~III pairs the ground-truth anomaly description with three adversarial
distractors. We use GPT-OSS-120B \citep{gptoss2025} as the distractor
generator, conditioned on the ground-truth (L1, L2, L3) annotation and on
the Video-HOCA taxonomy in Table~\ref{tab:taxonomy-def}. The prompt
deliberately samples one distractor from the same L1 branch (a different L3
under the same Ontological--Causal branch), one from the opposite L1 branch,
and one ``adversarial'' distractor that sounds physical but contradicts the
actual video event. The released review tool
(Figure~\ref{fig:anno-tool}) is used by human reviewers to keep, swap, or
edit each generated distractor before the multi-choice item is finalized.

{\small
\begin{verbatim}
You are an expert physics question generator for a Video-HOCA multi-choice
question.
Your task is to translate the provided ground truth (Chinese) into English
and then generate 3 plausible "distractor" descriptions.

### Taxonomy Definitions
{VIDEO_HOCA_TAXONOMY_DEFINITIONS}        // see Table tab:taxonomy-def for
                                         // the full L2/L3 hierarchy

### Input (Ground Truth)
- Scene Setup (Chinese):     "{scene_zh}"
- Observed Anomaly (Chinese):"{phenomenon_zh}"
- Classification:            {l1} -> {l2} -> {l3}

### Task 1: Translation
1. Translate Scene: render the Scene Setup into natural, neutral English
   (e.g., "A ceramic cup sits on a wooden table.").
2. Translate Anomaly (correct option): render the Observed Anomaly into
   English. This is the correct answer.
   - Constraint: must be a factual visual description of the event
     (e.g., "The cup floats upwards spontaneously.").

### Task 2: Distractor Generation (3 false options)
- Distractor 1 (Hard, same L1 branch): generate a description of a
  different anomaly type that belongs to the SAME L1 branch
  (Ontological or Causal).
   - Logic: if GT is Gravity (Causal), generate a Collision description
     for the same objects in the scene.
   - Example: "The cup passes through the table directly."
- Distractor 2 (Medium, different L1 branch): generate a description from
  the OTHER L1 branch.
   - Logic: if GT is Gravity (Causal), generate a Morphing description
     (Ontological).
   - Example: "The cup suddenly turns into a bird."
- Distractor 3 (Adversarial, logical/normalcy): generate a description
  that sounds physical but is either
    (a) a slightly wrong version of the actual event (e.g., "The cup
        falls sideways"); or
    (b) a non-anomalous outcome that contradicts the video (e.g., "The
        cup shatters upon impact" when it actually floated).
   - Goal: a valid English sentence that clearly describes a different
     outcome than what happened in the video.

### Output (JSON)
{
  "scene_description_en":     "...",   // translated scene
  "correct_option":           "...",   // translated anomaly
  "distractor_1_hard":        "...",   // same L1, different L3
  "distractor_1_type":        "...",   // L3 category for distractor 1
  "distractor_2_medium":      "...",   // different L1
  "distractor_2_type":        "...",   // L3 category for distractor 2
  "distractor_3_adversarial": "..."    // adversarial description
}
\end{verbatim}
}

\subsection{Additional Diagnostic Probes}
\label{app:additional-probes}

Table~\ref{tab:additional-probes} reports three auxiliary probes:
(a)~temporal grounding, (b)~multi-anomaly detection, and (c)~frame-count
sensitivity. Multi-anomaly numbers in (b) use the original 64-frame
stress-test setting with Thinking mode where supported, so they are
reported as a stress test rather than part of the Instruct-only main
leaderboard.

\paragraph{Temporal grounding.}
The two models in Table~\ref{tab:additional-probes}(a) show the gap between
a general-purpose Video-LLM and a temporal-localization specialist.
Time-R1-7B~\citep{timer1_2025}, which is RL-trained for temporal
localization, reaches mIoU $30.3$ and R@0.5 $15.5$---more than $3\times$ the
mIoU of vanilla Qwen2.5-VL-7B ($9.0$). Targeted RL training therefore
helps anomaly-event localization noticeably, although absolute numbers
remain modest, leaving headroom for future temporal-grounding work on
physical anomalies.

\paragraph{Multi-anomaly detection.}
F1 in Table~\ref{tab:additional-probes}(b) sits in the $43$--$57$ range
while exact-match accuracy is very low ($1.7$--$15.0$). Identifying multiple
concurrent violations is therefore substantially harder than the
single-anomaly main tasks, even under 64-frame Thinking-mode input; the
multi-anomaly probe is included as a stress test rather than as a target
for the main leaderboard.

\paragraph{Multi-frame scaling.}
Table~\ref{tab:additional-probes}(c) uses the same deterministic
Instruct-mode decoding as the main table (temperature $=0$, no Thinking
mode); the only change is the frame budget, going from 16 to 64 uniformly
sampled frames. Most models slightly regress on the Overall column under
this change---Gemini-2.5-Flash $-3.5$, GLM-4.6V-106B-FP8 $-3.0$,
Qwen3-VL-32B $-0.2$---with only Qwen3-VL-8B essentially flat ($+0.1$). The
regression patterns differ across models: Gemini-2.5-Flash loses most on
Task~I plausibility (ACC $75.5 \to 70.1$, F1 $77.7 \to 70.4$) while
Task~II/III stay roughly flat, whereas GLM-4.6V-106B-FP8 keeps its
16-frame Task~I level but craters on attribution and recognition,
including Task~II Causal ($21.9 \to 2.3$) and Task~III Ontological
($24.9 \to 17.3$). Under Instruct decoding alone, more frames therefore
do not automatically help, and the failure mode is model-specific rather
than a single shared mechanism.

\begin{table}[t]
    \centering
    \caption{Additional diagnostic probes. (a) Temporal grounding reports mean
             IoU and recall at IoU $=0.5$. (b) Multi-anomaly detection reports
             F1 and exact-match accuracy under the original 64-frame stress-test
             setting. (c) Multi-frame evaluation comparison on Video-HOCA. Overall uses the Task-wise means and standard deviations computed from the 16f main-table models only; both 16f and 64f rows report 50 plus 10 times the average task-wise z-score.}
    \label{tab:additional-probes}
    \footnotesize
    \setlength{\tabcolsep}{3.5pt}
    \begin{minipage}{0.48\linewidth}
        \centering
        \textbf{(a) Temporal grounding}\\[2pt]
        \begin{tabular}{lcc}
            \toprule
            \textbf{Model} & \textbf{mIoU} & \textbf{R@0.5} \\
            \midrule
            Time-R1-7B\citep{timer1_2025} & 30.3 & 15.5 \\
            Qwen2.5-VL-7B & 9.0 & 6.0 \\
            \bottomrule
        \end{tabular}
    \end{minipage}
    \hfill
    \begin{minipage}{0.48\linewidth}
        \centering
        \textbf{(b) Multi-anomaly detection}\\[2pt]
        \begin{tabular}{lcc}
            \toprule
            \textbf{Model} & \textbf{F1} & \textbf{Acc.} \\
            \midrule
            InternVL-3.5-38B & 43.3 & 1.7 \\
            Qwen3-VL-32B & 57.0 & 12.5 \\
            GLM-4.6V-106B-FP8 & 47.0 & 15.0 \\
            \bottomrule
        \end{tabular}
    \end{minipage}

    \vspace{0.8em}

    \textbf{(c) Multi-frame scaling}\\[2pt]
    \resizebox{\linewidth}{!}{%
    \begin{tabular}{llccccccccc}
        \toprule
        \multirow{2}{*}{\textbf{Model}} & \multirow{2}{*}{\textbf{Input}} & \multicolumn{2}{c}{\textbf{Task~I}} & \multicolumn{2}{c}{\textbf{Task~II}} & \multicolumn{2}{c}{\textbf{Task~III}} & \multicolumn{2}{c}{\textbf{Task~IV}} & \multirow{2}{*}{\textbf{Overall}} \\
        \cmidrule(lr){3-4}\cmidrule(lr){5-6}\cmidrule(lr){7-8}\cmidrule(lr){9-10}
        & & \textbf{ACC} & \textbf{F1} & \textbf{Ontological} & \textbf{Causal} & \textbf{Ontological} & \textbf{Causal} & \textbf{Ontological} & \textbf{Causal} & \\
        \midrule
        \multirow{2}{*}{Gemini-2.5-Flash} & 16f & 75.5 & 77.7 & 42.8 & 59.9 & 82.8 & 73.0 & 73.6 & 65.6 & 55.1 \\
         & 64f & 70.1 & 70.4 & 40.4 & 59.3 & 82.9 & 71.7 & 76.2 & 63.4 & 51.6 (\textcolor{red}{-3.5}) \\
        \midrule
        \multirow{2}{*}{GLM-4.6V-106B} & 16f & 69.8 & 76.2 & 31.5 & 21.9 & 24.9 & 18.0 & 71.0 & 58.9 & 41.8 \\
         & 64f & 69.2 & 75.8 & 28.2 & 2.3 & 17.3 & 16.7 & 71.9 & 59.5 & 38.8 (\textcolor{red}{-3.0}) \\
        \midrule
        \multirow{2}{*}{Qwen3-VL-8B} & 16f & 75.4 & 78.2 & 39.8 & 49.7 & 83.7 & 76.6 & 64.6 & 63.7 & 53.5 \\
         & 64f & 75.5 & 78.1 & 39.7 & 49.7 & 83.4 & 77.0 & 65.3 & 64.5 & 53.6 (\textcolor{green}{+0.1}) \\
        \midrule
        \multirow{2}{*}{Qwen3-VL-32B} & 16f & 79.5 & 82.3 & 46.0 & 55.6 & 84.1 & 79.0 & 78.5 & 62.9 & 57.4 \\
         & 64f & 79.4 & 82.2 & 47.1 & 55.7 & 84.1 & 79.2 & 68.8 & 63.4 & 57.2 (\textcolor{red}{-0.2}) \\
        \bottomrule
    \end{tabular}%
    }
\end{table}

\subsection{Temporal Frame-Order Ablation}
\label{app:temporal-frame-ablation}

Table~\ref{tab:shixu-temporal-frame-ablation} reports the temporal frame-order
ablation referenced in \S\ref{sec:input-controls}. We compare three input
settings on Qwen3.5-122B-A10B-FP8: the original 16-frame uniformly sampled
input, a single middle frame, and the same 16 frames in randomly shuffled
order. The middle-frame setting hurts every reasoning-heavy column (Tasks~II,
III, IV both branches), confirming that single-frame perception is
insufficient. The shuffled-frame setting stays close to the original order on
Tasks~I and~III but clearly degrades on Task~IV Ontological, indicating that
the model relies more on multi-frame coverage than on precise frame order.

\begin{table}[t]
    \centering
    \caption{Temporal frame-order ablation on Video-HOCA. The experiment
             compares the original 16-frame input, a single middle-frame input,
             and a shuffled 16-frame input. Task~I reports ACC/F1 (\%); Task~II
             and Task~III report Ontological--Causal scores (\%); Task~IV reports
             rubric-based Ontological--Causal judge scores (0--100). Bold marks
             the best temporal setting in each column.}
    \label{tab:shixu-temporal-frame-ablation}
    \scriptsize
    \setlength{\tabcolsep}{3.0pt}
    \resizebox{\linewidth}{!}{%
    \begin{tabular}{lllcccccccc}
        \toprule
        \multirow{2}{*}{\textbf{Model}} & \multirow{2}{*}{\textbf{Temporal setting}} & \multirow{2}{*}{\textbf{Frames}} & \multicolumn{2}{c}{\textbf{Task~I}} & \multicolumn{2}{c}{\textbf{Task~II}} & \multicolumn{2}{c}{\textbf{Task~III}} & \multicolumn{2}{c}{\textbf{Task~IV}} \\
        \cmidrule(lr){4-5}\cmidrule(lr){6-7}\cmidrule(lr){8-9}\cmidrule(lr){10-11}
          & & & \textbf{ACC} & \textbf{F1} & \textbf{Onto.} & \textbf{Caus.} & \textbf{Onto.} & \textbf{Caus.} & \textbf{Onto.} & \textbf{Caus.} \\
        \midrule
        Qwen3.5-122B-A10B-FP8        & Original order & 16f & \textbf{88.0} & \textbf{87.5} & \textbf{45.0} & \textbf{41.0} & \textbf{85.9} & \textbf{79.2} & \textbf{73.1} & \textbf{61.4} \\
        Qwen3.5-122B-A10B-FP8        & Middle frame   & 1f  & 80.6 & 82.0 & 39.5 & 40.2 & 77.9 & 70.7 & 58.0 & 59.0 \\
        Qwen3.5-122B-A10B-FP8        & Shuffled order & 16f & 87.9 & 87.2 & 44.4 & 36.8 & 81.8 & 75.5 & 68.5 & 59.0 \\
        \bottomrule
    \end{tabular}}
\end{table}

\subsection{Full Static/Dynamic Results}
\label{app:static-dynamic-full}

Table~\ref{tab:static-dynamic-full} provides the per-model breakdown of the
static/dynamic analysis summarized in
Table~\ref{tab:static-dynamic-summary} and \S\ref{sec:static-vs-onto}. Each
of the 20 main-table models is evaluated separately on the $258$ static and
$372$ dynamic anomaly clips from \texttt{static\_dynamic\_labels.jsonl}.
The three model groups---open-weight dense ($<$7B and $\geq$7B), open-weight
MoE/100B-class FP8, and closed-source---make it possible to read off whether
the static-vs.-dynamic gap depends on model family or scale: every model
loses Task~I accuracy on dynamic clips, while Task~II/III gaps are
considerably more uneven (some MoE configurations even reverse direction on
Task~II), supporting our claim that the Ontological--Causal gap is not
reducible to temporal demand alone.

\footnotesize
\setlength{\tabcolsep}{4.5pt}
\begin{longtable}{llrrrrr}
    \caption{Static vs.\ dynamic subset analysis on Video-HOCA. The subset labels
             come from \texttt{static\_dynamic\_labels.jsonl}. Only labeled
             anomaly videos are included, so Task~I is reported as subset
             accuracy rather than F1. Task~II/III report exact-match accuracy
             (\%), and Task~IV reports the rubric-based judge score (0--100).}
    \label{tab:static-dynamic-full}\\
    \toprule
    \textbf{Model} & \textbf{Subset} & \textbf{Videos} & \textbf{Task~I} & \textbf{Task~II} & \textbf{Task~III} & \textbf{Task~IV} \\
    \midrule
    \endfirsthead
    \toprule
    \textbf{Model} & \textbf{Subset} & \textbf{Videos} & \textbf{Task~I} & \textbf{Task~II} & \textbf{Task~III} & \textbf{Task~IV} \\
    \midrule
    \endhead
    \bottomrule
    \endfoot
    \rowcolor{black!8}\multicolumn{7}{c}{\emph{Labeled videos: Static=258, Dynamic=372}} \\
    \midrule
    \rowcolor{black!8}\multicolumn{7}{c}{\emph{Open-weight dense models ($<$7B)}} \\
    InternVL-3.5-2B              & Static  & 258 & 48.1 & 41.1 & 65.5 & 49.9 \\
                                 & Dynamic & 372 & 27.2 & 39.8 & 57.0 & 41.4 \\
    Qwen3-VL-2B                  & Static  & 258 & 70.9 & 52.7 & 76.7 & 56.1 \\
                                 & Dynamic & 372 & 49.7 & 37.4 & 65.1 & 47.0 \\
    \midrule
    \rowcolor{black!8}\multicolumn{7}{c}{\emph{Open-weight dense models ($\geq$7B)}} \\
    InternVL-2.5-8B              & Static  & 258 & 45.3 & 2.4 & 13.8 & 2.7 \\
                                 & Dynamic & 372 & 32.8 & 8.1 & 12.1 & 1.2 \\
    Qwen2.5-VL-7B                & Static  & 258 & 62.4 & 44.5 & 74.4 & 60.4 \\
                                 & Dynamic & 372 & 41.4 & 33.2 & 56.5 & 48.6 \\
    InternVL-3.5-8B              & Static  & 258 & 54.7 & 40.9 & 76.5 & 55.0 \\
                                 & Dynamic & 372 & 30.4 & 31.6 & 66.0 & 46.6 \\
    Qwen3-VL-8B                  & Static  & 258 & 72.9 & 42.0 & 86.4 & 66.2 \\
                                 & Dynamic & 372 & 55.6 & 33.3 & 74.8 & 62.6 \\
    Qwen3.5-9B                   & Static  & 258 & 70.5 & 21.7 & 71.2 & 68.8 \\
                                 & Dynamic & 372 & 59.4 & 25.3 & 69.8 & 62.7 \\
    InternVL-3.5-14B             & Static  & 258 & 64.7 & 42.7 & 79.6 & 61.4 \\
                                 & Dynamic & 372 & 44.4 & 39.2 & 68.9 & 57.2 \\
    Qwen3.5-27B                  & Static  & 258 & 85.3 & 45.7 & 89.4 & 72.1 \\
                                 & Dynamic & 372 & 79.0 & 48.8 & 86.0 & 68.7 \\
    Qwen3-VL-32B                 & Static  & 258 & 73.6 & 45.3 & 85.8 & 74.8 \\
                                 & Dynamic & 372 & 57.0 & 36.4 & 77.9 & 65.3 \\
    InternVL-3.5-38B             & Static  & 258 & 74.8 & 38.6 & 81.8 & 67.5 \\
                                 & Dynamic & 372 & 61.8 & 34.4 & 74.9 & 56.7 \\
    \midrule
    \rowcolor{black!8}\multicolumn{7}{c}{\emph{Open-weight MoE and 100B-class FP8 deployments}} \\
    GLM-4.6V-9B-Flash            & Static  & 258 & 45.0 & 40.4 & 78.6 & 61.1 \\
                                 & Dynamic & 372 & 20.7 & 32.1 & 66.9 & 61.5 \\
    InternVL-3.5-30B-A3B         & Static  & 258 & 59.3 & 39.0 & 83.1 & 60.0 \\
                                 & Dynamic & 372 & 37.1 & 36.7 & 70.7 & 50.6 \\
    Qwen3-VL-30B-A3B             & Static  & 258 & 74.8 & 41.3 & 83.3 & 76.8 \\
                                 & Dynamic & 372 & 61.3 & 33.6 & 72.4 & 62.8 \\
    Qwen3.5-35B-A3B              & Static  & 258 & 77.9 & 1.1 & 33.0 & 70.3 \\
                                 & Dynamic & 372 & 66.7 & 7.7 & 24.9 & 64.8 \\
    GLM-4.6V-106B-FP8            & Static  & 258 & 53.9 & 15.7 & 22.8 & 64.6 \\
                                 & Dynamic & 372 & 34.4 & 18.1 & 19.8 & 63.3 \\
    Qwen3.5-122B-A10B-FP8        & Static  & 258 & 94.6 & 35.1 & 85.2 & 64.4 \\
                                 & Dynamic & 372 & 90.1 & 30.4 & 79.7 & 67.3 \\
    \midrule
    \rowcolor{black!8}\multicolumn{7}{c}{\emph{Closed-source systems}} \\
    GPT-4o                       & Static  & 258 & 74.8 & 22.5 & 58.9 & 67.3 \\
                                 & Dynamic & 372 & 65.9 & 21.4 & 57.3 & 64.4 \\
    Gemini-2.5-Flash             & Static  & 258 & 74.4 & 39.8 & 78.3 & 70.1 \\
                                 & Dynamic & 372 & 59.1 & 43.9 & 75.7 & 67.9 \\
    Gemini-3-Flash               & Static  & 258 & 82.9 & 40.9 & 85.9 & 72.7 \\
                                 & Dynamic & 372 & 73.1 & 47.7 & 80.6 & 70.3 \\
\end{longtable}
\normalsize

\subsection{Task-IV Judge Rubric}
\label{app:judge}

Task~IV is \emph{not} evaluated by free-form commonsense judging. For each
question, we ship a structured reference answer that enumerates (i) scene
description, (ii) anomaly identification, (iii) anomaly type (Ontological vs.\
Causal), and (iv) physical explanation. The judge prompt asks GPT-OSS-120B
\citep{gptoss2025} to score the model's response against these four components
using a fixed rubric on a 0--100 scale. Scores are assigned against the
reference answer, not on free-form plausibility.

We validate this protocol in two ways: (i) on 100 sampled Task-IV responses
from representative models, human scoring vs.\ GPT-OSS-120B yields Pearson
$0.8957$ and Spearman $0.8914$; (ii) human-authored reference answers score
$81.4$ under GPT-OSS-120B and $84.9$ under DeepSeek-V4-Flash.

\paragraph{Field-name mapping.} The verbatim prompt below uses
\texttt{scene\_score}, \texttt{anomaly\_score}, \texttt{process\_score}, and
\texttt{reasoning\_score} as the four rubric fields, which correspond
respectively to the main-text components \emph{scene description},
\emph{anomaly identification}, \emph{anomaly type}, and \emph{physical
explanation}.

\paragraph{Verbatim judge prompt.} The exact prompt template sent to the
judge model (with reference answer \texttt{gt\_text} and model output
\texttt{prediction\_text}) is:

{\small
\begin{verbatim}
You are an expert physics professor and visual reasoning judge. Your task
is to evaluate a model's physical reasoning performance based on a video's
ground-truth (GT) and the model's prediction.

### Evaluation Rubric (Total 100 pts)

**Significant guideline: physical and semantic equivalence.**
The evaluation should prioritize accurate identification of the
underlying physical violation over exact linguistic matching.
- Granularity and summarization: GT references are often highly
  detailed. Predictions that provide a concise summary capturing the
  core physical event should be rewarded. If the prediction identifies
  the primary entity and the central action, it should receive full
  credit for scene and anomaly identification. Ignore missing
  environmental details (background, lighting, minor textures) and do
  not penalize brevity. Think like an intelligent evaluator, not a
  keyword matcher.
- Object-class flexibility: using functionally related terms (e.g.,
  "paintbrush" instead of "pencil") is acceptable if the visual
  features are ambiguous or similar.
- Physical essence vs. academic terms: focus on common sense and
  intuition. Full credit requires correctly identifying the violated
  physical law/common sense AND describing the normal expectation
  (e.g., "It violates the law of inertia because the object moves on
  its own; normally, it should stay still."). Academic terminology is
  NOT required for full marks.
   - Note on generic terms: if the model only says "it violates
     physical common sense" without explaining the specific phenomenon
     (e.g., "object should stay still"), deduct partial points
     (-10 pts). The explanation must relate to the specific event.

**Rubric summary**
1. Scene (25 pts):
   - Entity identification (10 pts): award 10 pts if the primary
     subject (e.g., "lamp", "cat", "dog") is correctly identified,
     even if specific sub-parts (e.g., "lamp shade") are omitted.
   - Action/behavior (10 pts): award 10 pts if the core motion or
     event is captured. Be lenient: generic terms like "moving" or
     "changing" are acceptable summaries for more specific actions
     like "rotating" or "morphing".
   - Outcome/context (5 pts): award 5 pts for describing the final
     state or environmental details.
   - Full-credit logic: if the prediction is a concise and accurate
     summary of "who did what", award the full 25 pts. Do not
     penalize for brevity compared to a highly detailed GT.
2. Anomaly (25 pts): physical essence of the violation. If the
   prediction identifies that a physical impossibility occurred to the
   correct object, award at least 15-20 pts even if the description is
   simple (e.g., "moving by itself" vs. "rotating independently").
3. Process (15 pts): exact label match (Causal/Ontological).
4. Reasoning (35 pts):
   - Full credit: correctly explains what scientific common
     sense/physical law was violated AND describes what the normal
     behavior of the event should be. Academic terms or common-sense
     descriptions are BOTH acceptable.
   - Partial deduction (-10): if the explanation is tautological
     (e.g., just says "violates physics" without explaining what
     behavior is wrong and what is normal).
   - Note: do NOT penalize for "imprecise terminology" if the logic
     is sound.

### Few-Shot Examples for Grading

Example 1: Perfect alignment with full reasoning (full marks)
- Ground-truth:
  { "scene": "A cat sits on a windowsill. Its tail extends to an
              extreme length.",
    "anomaly": "The cat's tail extends to an impossible length.",
    "process": "Ontological",
    "reasoning": "Violates biological regularities. Normally, a cat's
                  tail has a fixed skeletal structure and cannot grow
                  indefinitely." }
- Model prediction:
  "A cat sits by the window. Its tail is incredibly long, like a
   snake. This is Ontological. It violates biological common sense
   because a tail cannot grow to such an extreme length. Normally, a
   cat's tail has a fixed structure and fixed length."
- Judge output:
  { "breakdown": { "scene_score": 25, "anomaly_score": 25,
                   "process_score": 15, "reasoning_score": 35 },
    "justification": "The model correctly captures the scene and tail
                      anomaly. Its reasoning identifies the biological
                      violation (15 pts) and explicitly describes the
                      normal state (20 pts), providing a complete
                      explanation." }

Example 2: Generic violation and missing normal state (low reasoning)
- Ground-truth:
  { "scene": "A baseball bat hits a glass bottle on a wooden floor.
              The bottle remains stationary after impact.",
    "anomaly": "The bottle remains still while shards fly outwards.",
    "process": "Causal",
    "reasoning": "Violates momentum conservation. Normally, the bottle
                  should move due to impact." }
- Model prediction:
  "A bat hits a bottle. The bottle doesn't move. Causal. This violates
   common sense."
- Judge output:
  { "breakdown": { "scene_score": 25, "anomaly_score": 25,
                   "process_score": 15, "reasoning_score": 5 },
    "justification": "Scene and anomaly are correct. For reasoning, it
                      only provides a generic 'violates common sense'
                      without event-specific violation content (5/15)
                      and fails to describe what should have happened
                      normally (0/20). Total reasoning: 5/35." }

Example 3: Precise identification but flawed reasoning (zero reasoning)
- Ground-truth:
  { "scene": "An apple on a table rotates around its axis without
              contact.",
    "anomaly": "Apple spins spontaneously without external force.",
    "process": "Ontological",
    "reasoning": "Violates the conservation of angular momentum.
                  Normally, a stationary object remains still unless
                  acted upon by a torque." }
- Model prediction:
  "The apple starts spinning by itself. Ontological. It violates
   Gravity."
- Judge output:
  { "breakdown": { "scene_score": 25, "anomaly_score": 25,
                   "process_score": 15, "reasoning_score": 0 },
    "justification": "Reasoning score: 0/35. Citing 'Gravity' as the
                      reason for rotation is physically incorrect. It
                      fails BOTH the violation identification and the
                      normal-state components because the premise is
                      flawed." }

Example 4: Label mismatch but perfect reasoning (reasoning full)
- Ground-truth:
  { "scene": "Cat food in a bowl increases while being eaten.",
    "anomaly": "Cat food quantity increases during consumption.",
    "process": "Causal",
    "reasoning": "Violates conservation of mass. Normally, food
                  quantity should decrease when eaten." }
- Model prediction:
  "A cat is eating. The food multiplies. This is Ontological. It
   violates the law of conservation of matter because the food
   quantity increases during the eating process. Normally, the amount
   of food should decrease as it is consumed."
- Judge output:
  { "breakdown": { "scene_score": 25, "anomaly_score": 25,
                   "process_score": 0, "reasoning_score": 35 },
    "justification": "Process: 0 (mismatch). Reasoning: 35/35. It
                      correctly identifies the violation (15/15) and
                      describes the normal expectation (20/20) in
                      separate, clear components." }

Example 5: Partial scene and generic anomaly (partial marks)
- Ground-truth:
  { "scene": "A table lamp's shade rotates independently while the
              base stays still.",
    "anomaly": "Lampshade rotates without external force.",
    "process": "Causal",
    "reasoning": "Violates inertia. Normally a shade doesn't rotate
                  on its own." }
- Model prediction:
  "A lamp moves around. Ontological. Violates common sense, objects
   should stay still."
- Judge output:
  { "breakdown": { "scene_score": 20, "anomaly_score": 15,
                   "process_score": 0, "reasoning_score": 10 },
    "justification": "Scene: 20/25. Correctly identifies entity
                      'lamp' (10) and action 'moves' (10), but misses
                      specific details. Anomaly: 15/25. Captures the
                      essence of self-motion but is very generic.
                      Process: 0 (mismatch). Reasoning: 10/35.
                      Identifies the violation generally (5/15) but
                      fails to describe the specific normal state of
                      a lamp (5/20), getting only partial credit." }

### Input Data
- Ground-truth reference:
{gt_text}

- Model's prediction:
{prediction_text}

### Output Format
Provide your evaluation in JSON. You may include reasoning before the
JSON block, but the final response must contain a valid JSON object
wrapped in a ```json ... ``` code block:

{
  "breakdown": {
    "scene_score":     0,
    "anomaly_score":   0,
    "process_score":   0,
    "reasoning_score": 0
  },
  "justification": "Detailed explanation for each score point based on
                    the rubric."
}
\end{verbatim}
}

\subsection{Ranking Robustness under Scoring Choices}
\label{app:aggregation}

Table~\ref{tab:ranking-sensitivity} compares rankings under two Task-IV judges
and two aggregation rules. Switching the judge from GPT-OSS to DeepSeek barely
changes the z-score ranking. Changing the aggregation rule from z-score averaging
to a raw weighted average causes only local swaps, mostly among mid-ranked
models; the top two entries remain unchanged.

\begin{table}[t]
    \centering
    \caption{Ranking robustness under alternative Task~IV judges and aggregation
             rules. Each cell reports score (rank); arrows show rank changes
             relative to the GPT-OSS z-score setting. The z-score variants
             standardize Task~I F1, count-weighted Task~II/III macro-F1, and the
             selected Task~IV judge score over the same main-table models. The
             weighted variants use raw scores with weights 0.2/0.2/0.2/0.4 for
             Tasks~I--IV.}
    \label{tab:ranking-sensitivity}
    \scriptsize
    \setlength{\tabcolsep}{3.2pt}
    \resizebox{\linewidth}{!}{%
    \begin{tabular}{lcccc}
        \toprule
        \textbf{Model} & \textbf{GPT-OSS z-score} & \textbf{DeepSeek z-score} & \textbf{GPT-OSS weighted} & \textbf{DeepSeek weighted} \\
        \midrule
        Qwen3.5-27B             & 61.2 (1) & 61.3 (1) $\cdot$ & 74.4 (1) $\cdot$ & 73.4 (1) $\cdot$ \\
        Gemini-3-Flash          & 60.2 (2) & 60.6 (2) $\cdot$ & 73.5 (2) $\cdot$ & 73.3 (2) $\cdot$ \\
        Qwen3.5-122B-A10B-FP8   & 57.6 (3) & 57.5 (3) $\cdot$ & 68.9 (5) $\downarrow$2 & 67.6 (5) $\downarrow$2 \\
        Qwen3-VL-32B            & 57.4 (4) & 57.4 (4) $\cdot$ & 70.7 (3) $\uparrow$1 & 69.4 (3) $\uparrow$1 \\
        Qwen3-VL-30B-A3B        & 55.6 (5) & 55.2 (6) $\downarrow$1 & 68.7 (6) $\downarrow$1 & 66.5 (6) $\downarrow$1 \\
        Gemini-2.5-Flash        & 55.1 (6) & 55.7 (5) $\uparrow$1 & 69.0 (4) $\uparrow$2 & 69.3 (4) $\uparrow$2 \\
        Qwen3-VL-8B             & 53.5 (7) & 53.4 (7) $\cdot$ & 66.3 (7) $\cdot$ & 64.9 (7) $\cdot$ \\
        InternVL-3.5-14B        & 52.8 (8) & 52.8 (8) $\cdot$ & 64.1 (9) $\downarrow$1 & 63.0 (9) $\downarrow$1 \\
        InternVL-3.5-38B        & 52.5 (9) & 52.5 (9) $\cdot$ & 64.6 (8) $\uparrow$1 & 63.5 (8) $\uparrow$1 \\
        Qwen3.5-9B              & 51.9 (10) & 51.4 (10) $\cdot$ & 63.1 (10) $\cdot$ & 60.6 (11) $\downarrow$1 \\
        InternVL-3.5-30B-A3B    & 50.3 (11) & 50.4 (11) $\cdot$ & 61.4 (12) $\downarrow$1 & 60.6 (12) $\downarrow$1 \\
        GLM-4.6V-9B-Flash       & 49.4 (12) & 49.4 (12) $\cdot$ & 62.2 (11) $\uparrow$1 & 61.1 (10) $\uparrow$2 \\
        Qwen2.5-VL-7B           & 49.0 (13) & 48.7 (13) $\cdot$ & 58.5 (13) $\cdot$ & 56.6 (14) $\downarrow$1 \\
        InternVL-3.5-8B         & 48.6 (14) & 48.1 (14) $\cdot$ & 58.2 (15) $\downarrow$1 & 56.0 (15) $\downarrow$1 \\
        Qwen3-VL-2B             & 48.0 (15) & 47.5 (15) $\cdot$ & 57.6 (16) $\downarrow$1 & 55.2 (16) $\downarrow$1 \\
        GPT-4o                  & 46.5 (16) & 47.1 (16) $\cdot$ & 58.4 (14) $\uparrow$2 & 58.8 (13) $\uparrow$3 \\
        Qwen3.5-35B-A3B         & 43.8 (17) & 43.9 (17) $\cdot$ & 51.3 (18) $\downarrow$1 & 50.5 (18) $\downarrow$1 \\
        InternVL-3.5-2B         & 43.3 (18) & 43.2 (18) $\cdot$ & 52.1 (17) $\uparrow$1 & 50.6 (17) $\uparrow$1 \\
        GLM-4.6V-106B-FP8       & 41.8 (19) & 42.0 (19) $\cdot$ & 50.1 (19) $\cdot$ & 49.6 (19) $\cdot$ \\
        InternVL-2.5-8B         & 21.6 (20) & 22.0 (20) $\cdot$ & 17.7 (20) $\cdot$ & 17.5 (20) $\cdot$ \\
        \bottomrule
    \end{tabular}}
\end{table}

\subsection{Artifact and Release Details}
\label{app:artifact-release}

All open-weight models were evaluated locally on a single node with
$8\times$ NVIDIA H20 96\,GB GPUs using a unified vLLM-based inference stack.
Closed-source models (GPT-4o, Gemini-2.5-Flash, Gemini-3-Flash) were
evaluated through their public APIs available at the time of submission.

The anonymized artifact package released with the paper contains the
\emph{benchmark} and the \emph{evaluation code} required to reproduce
Table~\ref{tab:main}: (i)~the Video-HOCA dataset, comprising videos (or
URLs with reconstruction scripts for clips that cannot be redistributed
directly), per-clip taxonomy labels, the four-task QA pairs, and split
files for the static/dynamic and temporal-grounding subsets;
(ii)~the unified vLLM-based evaluation harness with the prompt templates
and decoding configurations used in the paper;
(iii)~the Task-IV judge prompt and rubric templates from
Appendix~\ref{app:judge}; and (iv)~the metric and aggregation code that
turns model outputs into the Task~I/II/III/IV scores and the z-score and
weighted overall columns reported in
\S\ref{sec:aggregation} and Appendix~\ref{app:aggregation}. We do
\emph{not} ship raw per-model inference outputs or per-response judge
scores in this release; users reproduce those by running the provided
harness against their chosen model endpoints.

\paragraph{License and takedown policy.}
The full Video-HOCA release---dataset, annotations, taxonomy guideline,
Task-IV judge prompt and rubric, evaluation harness, and metric/aggregation
code---is distributed under the Creative Commons Attribution 4.0
International license (CC~BY~4.0). Real-world clips for the plausibility
task are derived from Panda-70M \citep{panda70m} via the curation pipeline
of \S\ref{sec:dataset}; generated anomaly clips are produced by, or
reused from, the 15 video generators listed in the diversity-and-coverage
paragraph, and each upstream generator's terms of use continue to apply
to the underlying visual content. The release manifest records the source
and original license or terms of use for every clip. We will honor
takedown requests from any upstream provider, video generator, or
original creator who objects to the inclusion of a particular clip, and
the corresponding record will be removed from the dataset and from the
manifest in the next release version.

\subsection{Additional Experimental Details}
\label{app:details}

All experiments are zero-shot with fixed prompts per task family. For
closed-source models we use the public APIs available at evaluation time
(OpenAI for GPT-4o \citep{gpt4o2024}, Google for Gemini-2.5-Flash
\citep{gemini25_2025}); for open-weight models we use a unified vLLM-based
inference stack. Task~II and Task~III predictions are parsed with a
single-choice extraction protocol, and Task~IV uses the rubric-based judge
protocol described in Appendix~\ref{app:judge}.

Main-table decoding uses deterministic Instruct-mode settings for all four
tasks: temperature $=0$, no sampling, and \texttt{max\_new\_tokens}=1024. The
Thinking-vs.-Instruct controls in \S\ref{sec:thinking-vs-instruct} use an
8{,}192-token output budget for the aligned Instruct and Thinking settings. The
Qwen3-VL and Qwen3.5 Thinking settings use temperature $=1.0$, top-$p=0.95$,
top-$k=20$; GLM-4.6V-106B uses temperature $=0.8$ and top-$p=0.6$.

Main-table video inputs use 16 uniformly sampled frames. The temporal-grounding
probe uses 200 interval-prediction QA pairs (covering 59 newly annotated
clips and 141 single-anomaly clips reused from the main set with added
temporal-interval labels) and reports mIoU and R@0.5. The multi-anomaly probe
uses 120 clips with concurrent anomalies and 64-frame input, reporting F1
and exact-match accuracy. The multi-frame scaling
probe compares 16-frame and 64-frame inputs on the four main tasks. The
full-video / middle-frame / shuffled-frame controls in
\S\ref{sec:input-controls} use 16 uniformly sampled frames, a single middle
frame, and 16 frames with their order randomly permuted, respectively.

\clearpage
% NeurIPS Paper Checklist is for review/submission; omit for arXiv preprint.
% \input{checklist}

\end{document}